\PassOptionsToPackage{table}{xcolor}
\documentclass[twocolumn]{fairmeta}

\usepackage{amsmath,amssymb}
\usepackage{colortbl}
\usepackage{enumitem}
\usepackage{fontawesome}
\usepackage{pifont}
\usepackage{tikz}
\usetikzlibrary{positioning,fit,backgrounds,calc,arrows.meta}
\usepackage{pgfplots}
\pgfplotsset{compat=1.16}
\definecolor{accent}{HTML}{0064E0}       % Meta blue — agent, primary
\definecolor{slateblue}{HTML}{4A6FA5}    % Slate blue — tools, code
\definecolor{tealgreen}{HTML}{2A9D8F}    % Teal green — success/correct
\definecolor{terra}{HTML}{C44E52}        % Terracotta — error/wrong
\definecolor{inputbg}{HTML}{D6E4F0}      % Light blue — question/input bg
\definecolor{panelborder}{HTML}{8899A6}  % Panel borders

\newcommand{\ours}{R3D}
\newcommand{\ntbAssetDir}{figures/v2}

\title{\ours{}: Quantitative 3D Spatial Reasoning for Egocentric Wearables}

\author{Maxwell Horton}
\author{Wei Lu}
\author{Quan Tran}
\author{Yury Astashonok}
\author{Kirmani Ahmed}
\author{Babak Damavandi}
\author{Anuj Kumar}
\author{Xiao Zhang}
\author{Seungwhan Moon}

\affiliation{Meta Reality Labs}

\metadata[\faGithub~Code]{\url{https://github.com/facebookresearch/r3d}}
\metadata[\faDatabase~Dataset]{\url{https://huggingface.co/datasets/facebook/r3d-bench}}
\metadata[\faEnvelope~Correspondence]{Maxwell Horton at \email{maxwellhorton@meta.com}}

\abstract{Quantitative 3D spatial reasoning from egocentric RGB-D video is an important capability for next-generation wearable assistants. Yet existing benchmarks do not reflect the challenges of handling (1) natural egocentric video, (2) posed RGB-D video inputs, and (3) challenging quantitative 3D spatial reasoning Q\&A. To fill this gap, we introduce \textbf{\ours{}-Bench} (\textbf{R}easoning in \textbf{3D}), a benchmark of 2,950 quantitative spatial reasoning questions across 15 types---spanning multiple-choice, distance-based, and volumetric reasoning questions---built on top of 57 egocentric video sequences from Aria Digital Twin~\citep{adt}. To set a strong baseline on this dataset, we introduce \textbf{\ours{}}, a model-agnostic spatial tool-calling framework. In contrast to existing approaches that directly embed 3D information into the model's input representation, \ours{} constructs a 3D scene from video using segmentation and depth-lifted object representations. It provides this information to an LLM through seven composable spatial tools. On \ours{}-Bench, \ours{} with Qwen3.6 35B-A3B outperforms all baselines $<$100B parameters, achieving 55.6\% mean relative accuracy, a substantial increase over the second-best approach (CuTR+Tools, 47.7\%). In the $\geq$100B parameter domain, our \ours{} Qwen3-VL 235B-A22B Thinking variant improves upon its base model by 7.7\% absolute, achieving 57.0\% accuracy and outperforming all open- and closed-source models except closed-source GPT-5.5 (59.1\%), with which it remains competitive. We publicly release both \ours{}-Bench and our full implementation.}

\begin{document}

\newcommand{\roboticon}{\raisebox{-1.5pt}{\begin{tikzpicture}[scale=1.1]
  \draw[line width=0.3pt] (0,0.16) -- (0,0.24);\fill[accent!80] (0,0.26) circle (0.025);
  \draw[rounded corners=0.8pt, fill=accent!15, draw=accent!70, line width=0.35pt] (-0.13,-0.08) rectangle (0.13,0.155);
  \fill[white] (-0.06,0.045) circle (0.032);\fill[black] (-0.06,0.045) circle (0.018);
  \fill[white] (0.06,0.045) circle (0.032);\fill[black] (0.06,0.045) circle (0.018);
  \draw[accent!70, line width=0.3pt] (-0.045,-0.035) .. controls (0,-0.06) .. (0.045,-0.035);
\end{tikzpicture}}}

\twocolumn[
\mymaketitle
\vskip 0.38cm
\centering
\resizebox{\textwidth}{!}{% Teaser figure. tikzpicture only; main.tex wraps it in \resizebox.
% Three panels left->right: Inputs -> R3D system -> Agentic conversation.

% The ICLR style file redefines \tiny as 6pt (vs 5pt in a standard 10pt
% article), which widens the object boxes below enough to push the last one out
% of panel 2. \objtiny pins them to the size this layout was designed around.
\providecommand{\objtiny}{\fontsize{5}{6}\selectfont}
\providecommand{\ntbAssetDir}{figures/v2}

% Friendly robot icon
\providecommand{\roboticon}{\raisebox{-1.5pt}{\begin{tikzpicture}[scale=1.1]
  \draw[line width=0.3pt] (0,0.16) -- (0,0.24);\fill[orange!80] (0,0.26) circle (0.025);
  \draw[rounded corners=0.8pt, fill=orange!15, draw=orange!70, line width=0.35pt] (-0.13,-0.08) rectangle (0.13,0.155);
  \fill[white] (-0.06,0.045) circle (0.032);\fill[black] (-0.06,0.045) circle (0.018);
  \fill[white] (0.06,0.045) circle (0.032);\fill[black] (0.06,0.045) circle (0.018);
  \draw[orange!70, line width=0.3pt] (-0.045,-0.035) .. controls (0,-0.06) .. (0.045,-0.035);
\end{tikzpicture}}}

\begin{tikzpicture}[
  font=\small,
  panellbl/.style={font=\sffamily\bfseries},
  panelbg/.style={draw=panelborder, rounded corners=5pt, thick},
  imgframe/.style={draw=panelborder!70, line width=0.3pt, inner sep=0pt},
  sysbox/.style={draw=panelborder, rounded corners=3pt, thick, fill=white, font=\sffamily\small, inner sep=4pt, minimum height=7mm},
  toolsbox/.style={draw=slateblue!70, fill=slateblue!5, rounded corners=4pt, thick, inner sep=4pt, align=left},
  objbox/.style={draw=panelborder!80, rounded corners=2pt, inner sep=2pt, align=center, font=\scriptsize, minimum width=1.26cm},
  ttbox/.style={draw=panelborder!50, fill=panelborder!3, rounded corners=3pt, font=\ttfamily\scriptsize, inner sep=4pt, align=left},
  qbubble/.style={draw=accent!30, fill=inputbg!50, rounded corners=4pt, inner sep=4pt, align=left, font=\scriptsize},
  flow/.style={-{Stealth[length=5pt]}, line width=1pt, panelborder!80},
  biflow/.style={{Stealth[length=5pt]}-{Stealth[length=5pt]}, line width=1pt, accent!80},
]
\def\ph{5.2cm}
\node[panelbg, minimum width=3.6cm, minimum height=\ph] (p1) at (0,0) {};
\node[panelbg, minimum width=8.9cm, minimum height=\ph, right=10mm of p1] (p2) {};
\node[panelbg, minimum width=6.1cm, minimum height=\ph, right=10mm of p2] (p3) {};
\node[panellbl, anchor=south west] at ([yshift=2pt]p1.north west) {R3D-Bench};
\node[panellbl, anchor=south west] at ([yshift=2pt]p2.north west) {R3D};
\coordinate (imgleft) at ($(p1.north west)+(1.05,-0.60)$);
\node[panellbl, font=\sffamily\scriptsize\bfseries, anchor=east] at ($(imgleft)+(-1mm,0)$) {RGB};
\node[imgframe, anchor=west] (r1) at (imgleft) {\includegraphics[width=0.75cm]{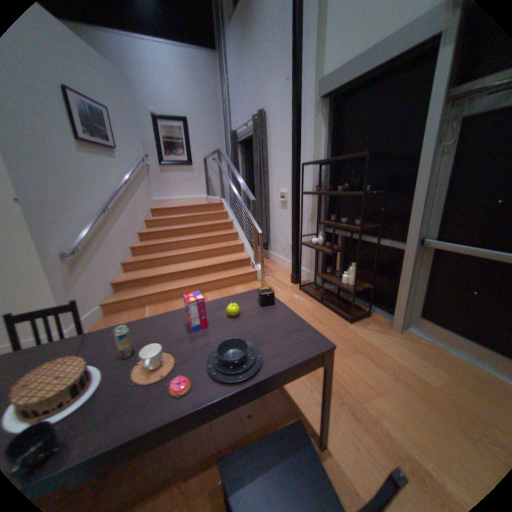}};
\node[imgframe, right=0.4mm of r1] (r2) {\includegraphics[width=0.75cm]{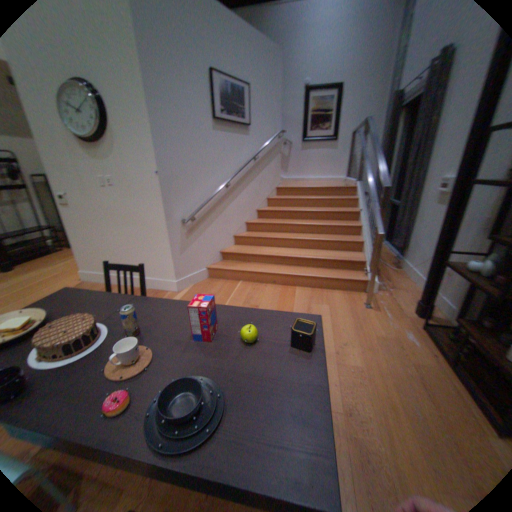}};
\node[imgframe, right=0.4mm of r2] (r3) {\includegraphics[width=0.75cm]{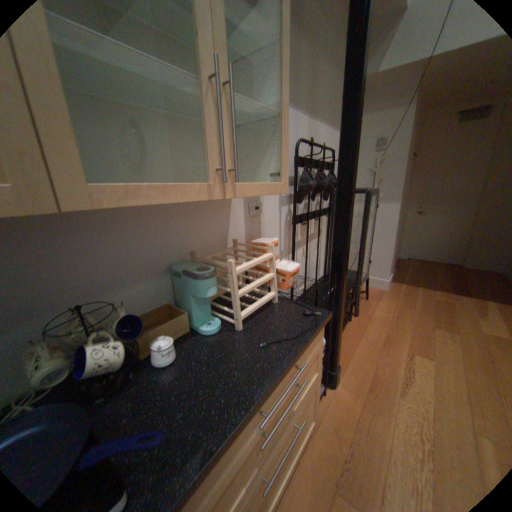}};
\node[panellbl, font=\sffamily\scriptsize\bfseries, anchor=east] at ($(imgleft)+(-1mm,-0.8)$) {Depth};
\node[imgframe, anchor=west] (d1) at ($(imgleft)+(0,-0.8)$) {\includegraphics[width=0.75cm]{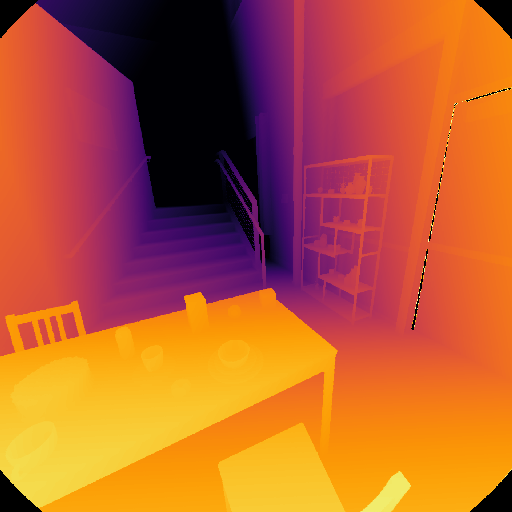}};
\node[imgframe, right=0.4mm of d1] (d2) {\includegraphics[width=0.75cm]{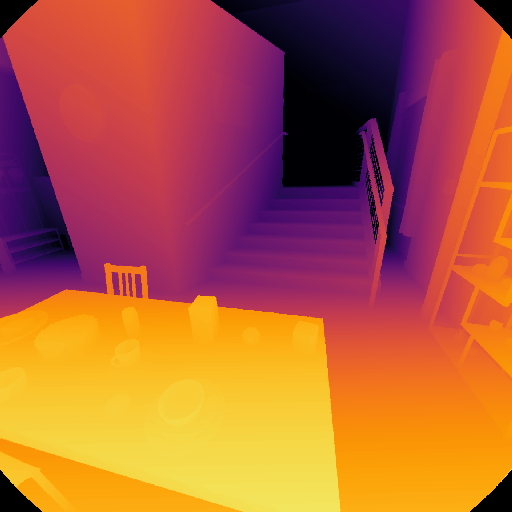}};
\node[imgframe, right=0.4mm of d2] (d3) {\includegraphics[width=0.75cm]{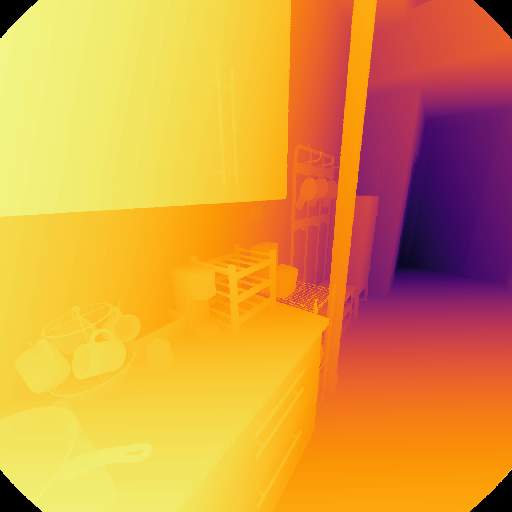}};
\node[anchor=east, font=\sffamily\scriptsize\bfseries] at ($(imgleft)+(-1mm,-1.55)$) {Pose};
\node[draw=panelborder!60, fill=panelborder!8, rounded corners=2pt, inner sep=3pt, anchor=west]
  (poseicon) at ($(imgleft)+(0,-1.55)$) {\scriptsize\faCamera};
\node[qbubble, text width=3.0cm, anchor=south west, font=\tiny] (question)
  at ($(p1.south west)+(0.2,0.15)$) {\textbf{Q:} ``After filling the black ceramic bowl with water and pouring into the mug until the mug is full, how many liters are left over in the black ceramic bowl?''};
\node[sysbox] (sam3) at ($(p2.north west)+(1.2,-0.6)$) {SAM3};
\node[toolsbox, anchor=east, font=\tiny] (tools) at ($(p2.north east)+(-0.15,-0.6)$) {%
  {\footnotesize\faWrench}\,\textbf{\scriptsize Tools}\\[1pt]
  \ttfamily list\_objects()\\
  \ttfamily get\_distance(id1,id2)\\
  \ttfamily get\_position(id)\\
  \ttfamily get\_my\_position()\\
  \ttfamily get\_distance\_from\_me(id)\\
  \ttfamily get\_object\_size(id)\\
  \ttfamily get\_object\_volume(id)};
\node[sysbox, fill=accent!8, draw=accent!70] (agent) at ($(sam3.east)!0.5!(tools.west)$) {\roboticon\,Agent};
\node[sysbox] (scene) at ($(sam3.south)+(0,-0.6)$) {Scene};
\draw[flow] (sam3) -- (scene);
\draw[flow] (scene.east) -| (agent.south);
\draw[biflow] (agent) -- (tools);
\node[objbox, anchor=north west] (obj1) at ($(p2.south west)+(0.3,2.05)$) {%
  \includegraphics[width=0.72cm]{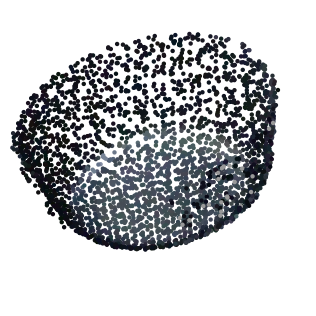}\\[-2pt]
  \textbf{\objtiny black ceramic bowl}\\
  \texttt{\objtiny pos\,(-2.1,0.8,2.9)}\\
  \texttt{\objtiny 14cm\!$\times$\!6cm\!$\times$\!14cm}};
\draw[flow] (scene.south) -- ++(0,-2mm) -| (obj1.north);
\node[objbox, anchor=north west] (obj2) at ($(obj1.north east)+(1mm,0)$) {%
  \includegraphics[width=0.72cm]{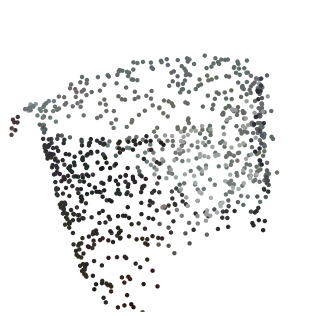}\\[-2pt]
  \textbf{\objtiny mug}\\
  \texttt{\objtiny pos\,(-1.7,0.8,3.0)}\\
  \texttt{\objtiny 9cm\!$\times$\!8cm\!$\times$\!11cm}};
\draw[flow] (scene.south) -- ++(0,-2mm) -| (obj2.north);
\node[objbox, anchor=north west] (obj3) at ($(obj2.north east)+(1mm,0)$) {%
  \includegraphics[width=0.72cm]{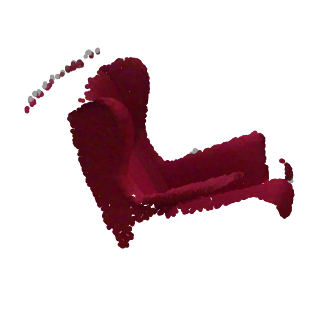}\\[-2pt]
  \textbf{\objtiny red wing chair}\\
  \texttt{\objtiny pos\,(-3.4,0.6,-0.4)}\\
  \texttt{\objtiny 106cm\!$\times$\!69cm\!$\times$\!151cm}};
\draw[flow] (scene.south) -- ++(0,-2mm) -| (obj3.north);
\node[objbox, anchor=north west] (obj4) at ($(obj3.north east)+(1mm,0)$) {%
  \includegraphics[width=0.72cm]{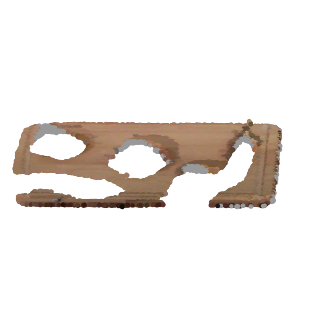}\\[-2pt]
  \textbf{\objtiny chopping board}\\
  \texttt{\objtiny pos\,(0.9,0.9,2.0)}\\
  \texttt{\objtiny 41cm\!$\times$\!4cm\!$\times$\!33cm}};
\draw[flow] (scene.south) -- ++(0,-2mm) -| (obj4.north);
\draw[flow] (r3.east) -- (sam3.west);
\coordinate (p1exit) at ($(p1.east)+(5mm,0)$);
\draw[flow] (d3.east) -- (d3.east -| p1exit) |- ([yshift=1mm]scene.west);
\draw[flow] (poseicon.east) -- (poseicon.east -| p1exit) |- ([yshift=-2mm]scene.west);
\node[qbubble, text width=5.5cm, anchor=south] (qbox) at ($(p3.north)+(0,-1mm)$) {%
  \textbf{Q:} ``After filling the black ceramic bowl with water and pouring into the mug until the mug is full, how many liters are left over in the black ceramic bowl?''};
\node[ttbox, text width=5.5cm, anchor=north, inner sep=4pt] (convo) at ($(qbox.south)+(0,-0.15)$) {%
  \roboticon\,\textbf{Agent:} list\_objects()\\
  \textcolor{gray}{$\rightarrow$ 31 objs: 19:black ceramic bowl, 18:mug\,\ldots}\\[1pt]
  \roboticon\,\textbf{Agent:} get\_object\_volume(19)\\
  \textcolor{gray}{$\rightarrow$ 0.554\,L}\\[1pt]
  \roboticon\,\textbf{Agent:} get\_object\_volume(18)\\
  \textcolor{gray}{$\rightarrow$ 0.294\,L}\\[1pt]
  \roboticon\,\textbf{Agent:} 0.554 $-$ 0.294 = \textbf{0.26\,L}\\[2pt]
  \textcolor{tealgreen}{\checkmark\,\textbf{Answer: 0.26\,L}}~~\textcolor{gray}{\scriptsize(GT 0.274, 5.1\% err)}};
\node[draw=terra!40, fill=terra!3, rounded corners=3pt, text width=5.5cm, inner sep=3pt,
      anchor=north, align=left, font=\scriptsize] (baselines) at ($(convo.south)+(0,-2mm)$) {%
  \textcolor{terra}{\ding{55}}~~GPT-5.5 (RGB): \textbf{0.15\,L}~~\textcolor{gray}{(45.3\% err)}\\
  \textcolor{terra}{\ding{55}}~~Gemini 3 Flash (RGB): \textbf{0.2\,L}~~\textcolor{gray}{(27.0\% err)}};
\end{tikzpicture}
}
\captionof{figure}{\textbf{Left:} \textbf{R3D-Bench} presents egocentric RGB-D frames, camera pose, and quantitative spatial reasoning questions. \textbf{Middle:} \textbf{\ours{}} uses SAM3 to segment objects in a video, then lifts them into a 3D scene. An LLM answers the question by calling spatial tools over the scene. \textbf{Right:} the LLM's tool-calling trace --- it identifies the black ceramic bowl and the mug, queries their volumes, and computes the water left over (5.1\% error), while GPT-5.5 and Gemini~3 Flash answer from RGB frames with 45\% and 27\% error.}
\label{fig:teaser}
\vskip 1em
]

%----------------------------------------------------------------------

\section{Introduction}
\label{sec:intro}

Wearable compute devices equipped with RGB-D images and real-time SLAM are rapidly maturing. Depth sensing and 6-DoF pose tracking are already standard on shipped augmented reality and mixed reality devices such as HoloLens~2~\citep{hololens2}, Quest~3~\citep{quest3}, and Apple Vision Pro~\citep{applevisionpro}. Research platforms like Project Aria~\citep{projectaria} are bringing SLAM-enabled wearables to a lightweight glasses form factor. A natural application of these sensors is equipping wearable AI systems with quantitative 3D spatial reasoning. Users wearing such devices will ask questions like ``If I fill my mug, how much liquid is left in my coffee can?'' or ``How far is the couch from the TV?'' Answering these questions requires quantitative spatial reasoning---producing outputs in absolute units (e.g. meters, liters) and applying reasoning on top of the outputs---rather than only evaluating qualitative relationships (e.g. ``above'', ``below'').

Existing 3D spatial reasoning benchmarks do not provide a realistic evaluation of the quantitative reasoning performance of wearable AI assistants. Many of them are purely image-based~\citep{spatialvlm,spatialrgpt}, or operate on pre-scanned 3D scenes with complete reconstructions~\citep{scanqa,sqa3d}, or use scanning walkthroughs with unnatural motion~\citep{vsibench,embodiedscan}, or focus only on qualitative rather than quantitative queries~\citep{openeqa}. None combine the three properties required to evaluate a next-generation wearable spatial AI assistant: (1)~\textbf{natural egocentric video}, (2)~\textbf{calibrated depth and camera pose}, and (3) \textbf{challenging quantitative 3D spatial reasoning Q\&A}.

To address this gap, we introduce \textbf{\ours{}-Bench} (Figure~\ref{fig:teaser}), a quantitative 3D spatial reasoning benchmark built on top of 57 video sequences from the Aria Digital Twin (ADT) dataset~\citep{adt}. ADT contains videos captured with Project Aria glasses during natural activities (cleaning, cooking, meal preparation, working, and decoration). Our Q\&A benchmark annotates these videos with 2,950 annotations across 15 question types in three categories: multiple choice (comparing two or more metric quantities), distance-based, and volume-based. Our dataset features accurate ground truth derived from detailed object annotations from ADT.

We find that many existing multimodal large language models (LLMs) perform poorly on \ours{}-Bench. Recent methods that incorporate depth information in a learned latent representation \citep{video3dllm,spatialrgpt} provide strong results on qualitative questions (e.g. ``the cup is to the left of the plate''), but do not generalize well to providing quantitative measurements (Table~\ref{tab:main_results}).

To provide a strong baseline on \ours{}-Bench, we develop \textbf{\ours{}} (Figure~\ref{fig:teaser}), a model-agnostic spatial tool-calling framework. Given egocentric video, depth, and pose, \ours{} segments the scene with SAM3~\citep{sam3} and lifts objects to 3D via depth back-projection. \ours{} then combines partial observations using multi-view consistency and KNN filtering to produce a point cloud. This point cloud is used to generate bounding boxes suitable for distance-based Q\&A. Additionally, meshes are generated with SAM3D~\citep{sam3d} and resized into metric space for answering volumetric questions. These bounding boxes and meshes are then queried by an LLM through seven spatial tools. Our method can be applied zero-shot to any LLM with tool calling capabilities.

On \ours{}-Bench, \ours{} with Qwen3.6 35B-A3B achieves 55.6\% mean relative accuracy, outperforming other 3D methods in the $<$100B parameter range (Table~\ref{tab:main_results}). When comparing larger models ($\geq$100B parameters), \ours{} with Qwen3-VL 235B-A22B Thinking achieves 57.0\%, outperforming closed-source  Gemini variants, and second only to GPT-5.5 (59.1\%), with which it remains competitive.

In summary, our contributions are:
\begin{itemize}[nosep]
  \item \textbf{\ours{}-Bench}: a challenging quantitative 3D spatial reasoning benchmark for egocentric wearables with 57 videos and 2,950 metric-space Q\&A annotations across 15 types.
  \item \textbf{\ours{}}: a model-agnostic spatial tool-calling framework that constructs a 3D scene from egocentric video, giving the best accuracy among depth-enabled methods and exceeding or competing with large-scale closed-source foundation models, with zero training FLOPs.
\end{itemize}

The rest of our paper is organized as follows: we discuss related works in Section~\ref{sec:related}. We describe construction of \ours{}-Bench in Section~\ref{sec:dataset}. We describe \ours{} in Section~\ref{sec:method}. Our main results appear in Section~\ref{sec:experiments}. We discuss limitations in Section~\ref{sec:limitations}. We conclude in Section~\ref{sec:conclusion}.

%----------------------------------------------------------------------
\section{Related Work}
\label{sec:related}

% Merged float: benchmark comparison table (a) + example visual inputs (b).
% These were two separate full-height floats in the two-column layout; in the
% single-column ICLR layout they fit side by side, which saves roughly a page.
\begin{figure*}[t]
\centering
\begin{subfigure}[b]{0.37\textwidth}
\centering
\scriptsize
\setlength{\tabcolsep}{2pt}
\begin{tabular}{@{}lcccc@{}}
\toprule
 & \rotatebox{70}{Nat.\ Egocentric} & \rotatebox{70}{Video} & \rotatebox{70}{Depth+Pose} & \rotatebox{70}{Quantitative Q\&A} \\
\midrule
ScanQA              & --          & --          & --          & --          \\
SQA3D               & --          & --          & --          & --          \\
EmbodiedScan       & --          & \checkmark  & \checkmark  & --          \\
SpatialVLM          & --          & --          & --          & \checkmark  \\
SpatialRGPT-Bench   & --          & --          & --          & \checkmark  \\
CA-VQA              & --          & --          & \checkmark  & \checkmark  \\
OpenEQA             & \checkmark  & \checkmark  & \checkmark  & --          \\
VSI-Bench           & --          & \checkmark  & --          & \checkmark  \\
OSI-Bench           & \checkmark  & \checkmark  & --          & \checkmark  \\
\midrule
\textbf{\ours{}-Bench} & \textbf{\checkmark} & \textbf{\checkmark} & \textbf{\checkmark} & \textbf{\checkmark} \\
\bottomrule
\end{tabular}
\caption{Comparison of \ours{}-Bench and existing benchmarks.}
\label{tab:benchmark_comparison}
\end{subfigure}
\hfill
\begin{subfigure}[b]{0.60\textwidth}
\centering
\setlength{\tabcolsep}{1pt}
\newcommand{\segfig}[1]{\includegraphics[width=0.135\textwidth]{\ntbAssetDir/seg/#1}}
\begin{tabular}{@{}m{0.4cm}m{0.135\textwidth}m{0.135\textwidth}m{0.135\textwidth}m{0.135\textwidth}@{}}
& \centering\scriptsize\textbf{wooden fork} & \centering\scriptsize\textbf{frying pan} & \centering\scriptsize\textbf{white utensil tray} & \centering\arraybackslash\scriptsize\textbf{black ceramic mug} \\[1pt]
\rotatebox{90}{\tiny\textbf{GT}} &
\segfig{col1_gt.png} & \segfig{col2_gt.png} & \segfig{col3_gt.png} & \segfig{col4_gt.png} \\[1pt]
\rotatebox{90}{\tiny\textbf{GT Crop}} &
\segfig{col1_gt_crop.png} & \segfig{col2_gt_crop.png} & \segfig{col3_gt_crop.png} & \segfig{col4_gt_crop.png} \\[2pt]
\rotatebox{90}{\tiny\textbf{SAM3}} &
\segfig{col1_sam3.png} & \segfig{col2_sam3.png} & \segfig{col3_sam3.png} & \segfig{col4_sam3.png} \\[1pt]
\rotatebox{90}{\tiny\textbf{SAM3 Crop}} &
\segfig{col1_sam3_crop.png} & \segfig{col2_sam3_crop.png} & \segfig{col3_sam3_crop.png} & \segfig{col4_sam3_crop.png} \\
\end{tabular}
\caption{Example visual inputs. Rows 1--2: ground-truth mask with zoomed crop. Rows 3--4: SAM3 mask on the same frame with zoomed crop.}
\label{fig:seg_comparison}
\end{subfigure}
\caption{Benchmark comparison and example visual inputs. The examples in (b) include challenging examples: a thin object (wooden fork), a distant object tracked with low IoU (frying pan), a track that locks onto the wrong object (white utensil tray), and an object SAM3 never tracks (black ceramic mug).}
\label{fig:bench_and_inputs}
\end{figure*}

\textbf{3D Q\&A Datasets} Many prior works explore Q\&A over 3D datasets. The primary difference with our work is that we explore the egocentric wearables use case. Several datasets pose semantic and relational questions assuming the availability of room scans rather than streaming egocentric video \citep{scanqa, sqa3d,3dllm,leo}. A few others introduce quantitative questions~\citep{vsibench,spatialrgpt,mmspatial_model}, but not in the RGB-D egocentric video setting. OSI-Bench~\citep{osibench} has accurate ground truth but captures with a LiDAR rig and does not provide aligned depth as an input. OpenEQA~\citep{openeqa} is complementary: an egocentric video QA benchmark focused on episodic memory rather than metric measurement. Figure~\ref{tab:benchmark_comparison} summarizes these differences.

\textbf{3D Q\&A Methods} Many previous works explore \emph{qualitative} 3D spatial reasoning. Video-3D LLM~\citep{video3dllm} encodes 3D position information directly into video token embeddings. Other works support open-vocabulary 3D queries by building 3D scenes \citep{conceptfusion,conceptgraphs,lerf,openscene}, or solve 3D Q\&A through inference techniques leveraging pre-scanned 3D scenes \citep{scanqa,sqa3d,3dllm,leo}. These works do not handle partially-observed objects from egocentric video or quantitative reasoning Q\&A.

Among works capable of quantitative spatial reasoning Q\&A, SpatialVLM~\citep{spatialvlm} and SpatialRGPT~\citep{spatialrgpt} address image-based Q\&A. SR-3D~\citep{sr3d} and MM-Spatial~\citep{mmspatial_model} incorporate depth and pose into spatial understanding on multi-view images.

A few works explore tool calling~\citep{toolformer,react,gorilla} for spatial reasoning \citep{spacetools,think3d,pyspatial,vadar}, but do not target calibrated egocentric RGB-D video from wearables. Concurrent to our work, RieMind~\citep{riemind} combines an LLM with a 3D scene graph and geometric tools similar to our method. However, it operates on ground truth 3D annotations rather than running perception itself, representing an upper bound rather than a deployable system. A few other concurrent works \citep{spatialclaw,sagent} investigate agentic spatial tool use over existing benchmarks. We differ in that our focus is on evaluation and deployment of a realistic wearable spatial AI benchmark.

%----------------------------------------------------------------------
\begin{table*}[!t]
\centering
\caption{The 15 question types in \ours{}-Bench with an example question, answer format, and question count. Each question has an average of 279.4 input frames (93.1 seconds at 3\,FPS). Bool = yes/no, Str = object name, Float = number in meters or liters.}
\label{tab:question_types}
\scriptsize
\setlength{\tabcolsep}{3pt}
\begin{tabular}{@{}lllrp{9.0cm}@{}}
\toprule
\textbf{Category} & \textbf{Type} & \textbf{Fmt} & \textbf{N} & \textbf{Example question} \\
\midrule
\multirow{5}{*}{\shortstack[l]{Multiple\\Choice}}
  & gap\_fit & Bool & 61 & ``Could I place the white chair in the gap between the refrigerator and the bookcase?'' \\
  & nearest\_from\_set & Str & 186 & ``Among the heat trivet, the bookcase, and the cereal box, which one am I closest to?'' \\
  & top\_higher & Bool & 217 & ``Is the top of the black wall shelf higher up than the top of the grey sofa?'' \\
  & which\_taller & Str & 225 & ``Between the coffee can and the DVD remote, which one is taller?'' \\
  & which\_longer\_dim & Str & 225 & ``Which has a greater longest dimension: the bird house toy or the kitchen island?'' \\
\midrule
\multirow{7}{*}{Distance}
  & global\_how\_far & Float & 226 & ``What is the distance between the white trash bin and the black coffee table, in meters?'' \\
  & global\_how\_far\_from\_me & Float & 225 & ``How many meters away is the coffee can from where I am?'' \\
  & global\_how\_long & Float & 226 & ``How big is the frying pan along its longest axis, in meters?'' \\
  & how\_much\_taller & Float & 224 & ``By how many meters is the black wall shelf taller than the large decorative box?'' \\
  & how\_much\_longer\_dim & Float & 226 & ``By how many meters does the large black frame exceed the small black ceramic dish in its longest dimension?'' \\
  & total\_fly\_distance & Float & 226 & ``What is the total straight-line distance from me to the center of the humidifier and then to the center of the blender, in meters?'' \\
  & total\_walk\_distance & Float & 225 & ``How far would I walk horizontally if I go to the center of the picture ledge first, then to the center of the blender, in meters?'' \\
\midrule
\multirow{3}{*}{Volume}
  & volume\_estimation & Float & 164 & ``How much liquid can the white vase hold, in liters?'' \\
  & pour\_leftover & Float & 147 & ``If I fill the jam jar completely with water and pour into the mug until it is full, how many liters remain in the jam jar?'' \\
  & pour\_room\_left & Float & 147 & ``If I fill the mug with water and pour it into the wooden bowl, how much room is left in the wooden bowl, in liters?'' \\
\midrule
\multicolumn{3}{@{}l}{Total} & 2,950 & \\
\bottomrule
\end{tabular}
\end{table*}

\section{\ours{}-Bench}
\label{sec:dataset}

Here we discuss the construction of \ours{}-Bench. In Section~\ref{sec:adt}, we discuss our choice of building annotations on top of Aria Digital Twin (ADT). In Section~\ref{sec:qa_construction}, we discuss the construction of Q\&A pairs. In Section~\ref{sec:dataset_characteristics}, we discuss characteristics of our challenging benchmark.

\subsection{Aria Digital Twin}
\label{sec:adt}

Our dataset is built on 57 sequences from Aria Digital Twin~\citep{adt}, captured with Project Aria glasses~\citep{projectaria}. This dataset contains 90--100 second RGB-D videos of everyday activities (cleaning, cooking, meal preparation, working, and decoration). Each frame has pose annotations and detailed object annotations. Our main motivations for using ADT are (1) natural egocentric video, (2) the presence of depth and pose information, and (3) rich annotations for quantitative Q\&A, including object meshes for volumetric estimation ground truth generation. In Figure~\ref{fig:seg_comparison}, we visualize a few images from Aria Digital Twin alongside segmentation masks produced by a state-of-the-art object detector, SAM3~\citep{sam3}. These examples highlight thin objects, distant objects, tracks that lock onto the wrong object, and objects SAM3 fails to track.

\begin{figure*}[!t]
\centering
\newlength{\dsfigh}\setlength{\dsfigh}{3.8cm}
\begin{subfigure}[t]{0.32\textwidth}
\vspace{0pt}
\centering
\parbox[t][\dsfigh][c]{\textwidth}{\centering\includegraphics[width=\textwidth,height=\dsfigh,keepaspectratio]{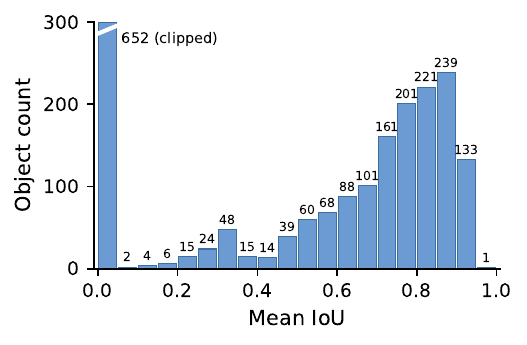}}
\caption{SAM3 IoU distribution.}
\label{fig:iou_histogram}
\end{subfigure}
\hfill
\begin{subfigure}[t]{0.32\textwidth}
\vspace{0pt}
\centering
\parbox[t][\dsfigh][c]{\textwidth}{\centering\includegraphics[width=0.75\textwidth,height=\dsfigh,keepaspectratio]{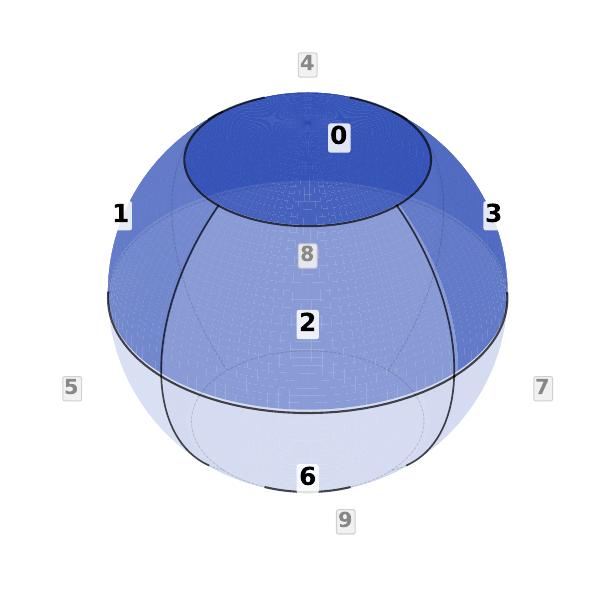}}
\caption{Viewpoint binning scheme.}
\label{fig:viewpoint_sphere}
\end{subfigure}
\hfill
\begin{subfigure}[t]{0.32\textwidth}
\vspace{0pt}
\centering
\parbox[t][\dsfigh][c]{\textwidth}{\centering\includegraphics[width=\textwidth,height=\dsfigh,keepaspectratio]{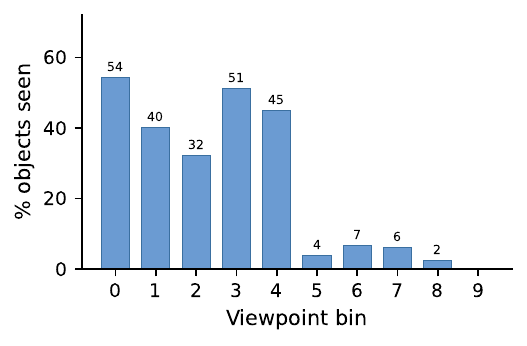}}
\caption{\% of objects seen from each bin.}
\label{fig:viewpoint_histogram}
\end{subfigure}
\caption{Dataset characteristics. (a) Mean SAM3 tracking IoU of all objects referenced by the 2,950 questions (some objects are never found, or misidentified). The first bar (mean IoU $<$ 0.05, clipped at 652) contains the 645 objects SAM3 never tracks and 7 tracked objects whose track mostly follows another object. (b) Viewpoint binning scheme used in (c), 10 bins of equal solid angle. (c) Percentage of objects seen from each binned viewpoint.}
\label{fig:dataset_stats}
\end{figure*}

\subsection{\ours{}-Bench Construction}
\label{sec:qa_construction}
\ours{}-Bench includes frame timestamps, questions, answers, and pointing annotations. We discuss generation of these elements here.

\textbf{Frame Timestamps}: An always-on AI assistant typically streams frames to cloud-based systems for analysis. Such systems cannot provide high-FPS image streams due to power and latency considerations. Thus, for consistent and fair evaluation, we first generate a list of valid frame timestamps at 3FPS. We use this list of timestamps to extract RGB-D frames (with fisheye distortion removed) and camera poses from ADT. We downsample images to 512x512, to simulate a low-power device streaming frames for cloud processing at acceptable resolution.

\textbf{Questions and Answers}: Next, we produce quantitative 3D spatial reasoning question and answer annotations for each sequence. First, we produce natural names for objects, mapping ADT names (e.g.\ ``AirPurifier\_1'') to natural names (e.g.\ ``air purifier''). We then define 15 question types organized into three categories (Table~\ref{tab:question_types}):
\begin{itemize}[nosep]
  \item \textbf{Multiple Choice:} e.g. ``Does the couch fit in the gap between the TV and the plant?''
  \item \textbf{Distance:} e.g. ``How much longer is the couch than the chair?''
  \item \textbf{Volume:} e.g. ``If I pour the flask into the mug, how much water is left?''
\end{itemize}
Each category contains 3--7 subtypes (Table~\ref{tab:question_types}). Each subtype has 2--4 natural-language templates for linguistic variety. Templates reference specific objects using their natural names and are populated from the ADT ground truth 3D annotations. Questions involving the user's position use the user's position at the end of the sequence. Ground truth answers are computed from ADT bounding boxes for distance-based questions, and functional mesh volumes for volume-based questions (see Appendix~\ref{app:gt_volume} for details).

\textbf{Pointing Annotations}: To disambiguate semantically similar objects (e.g. ``cup'' and ``glass''), we provide a ``pointing reference'' to each referred object in each question. This reference is a ground truth mask for the object the first time it appears in the video. We do not provide ground truth segmentations for any other timestamp.

\textbf{Filtering}: To ensure our benchmark focuses mainly on metric spatial Q\&A without contamination from object ambiguity or extremely heavily occluded objects, we keep objects with unique names that are static and clearly visible in several frames. We also filter out questions that are too obvious, such as ``Is the carrot taller than the refrigerator?'' See Appendix~\ref{app:qa_filtering} for details.
\subsection{Dataset Characteristics}
\label{sec:dataset_characteristics}
Our final dataset contains (1) natural egocentric video, (2) calibrated depth and camera pose, and (3) challenging quantitative 3D spatial reasoning Q\&A to provide a challenging benchmark for depth-enabled wearables. We highlight some of the challenges here:

\textbf{Visual Inputs:} As shown in Figure~\ref{fig:seg_comparison}, our dataset contains thin objects, small objects viewed from a distance, and objects that SAM3 tracks poorly or not at all. These all contribute to challenges in visually parsing objects. We show the distribution of SAM3 tracking IoU for each object in Figure~\ref{fig:iou_histogram}.

\textbf{Partial Observations:} Our final dataset exhibits challenging scenarios with partially-observed objects. In Figure~\ref{fig:viewpoint_sphere} and Figure~\ref{fig:viewpoint_histogram}, we visualize the distribution of viewpoints using an equal-area spherical binning scheme with 10 bins of equal solid angle. We note that 54.4\% of the 1,447 reconstructed objects are seen from above, but few objects are seen from below (2--7\% per bin, with 0\% of objects seen from directly below). This contrasts sharply with scan-based datasets~\citep{scannet,scannetpp} where deliberate scanning motions include low viewpoints with the camera pointing upwards.

\section{\ours{}: 3D Spatial Reasoning with Tools}
\label{sec:method}

To address the challenges posed by \ours{}-Bench, we develop \ours{}, a method for quantitative 3D spatial reasoning with tools. Motivated by recent works in agentic AI~\citep{react,toolformer}, we use tools to inject 3D information in LLMs. The core of our method involves aggregating 3D information from RGB-D images and pose in order to expose this information to an LLM capable of reasoning. We note that even 3D LLMs that operate in latent space still typically use object detectors to provide input signals to the LLMs~\citep{video3dllm}. Our approach pushes this concept further, providing rich 3D signals to the model in the form of tools.

Our overall approach is shown in Figure~\ref{fig:teaser}. Our method takes in RGB-D video, camera pose, and questions, and outputs answers through scene construction (Section~\ref{sec:scene_construction}) and multi-step tool calling (Section~\ref{sec:tool_calling}). Our method is model-agnostic, and can be integrated with any tool-calling-enabled reasoning model with no training required.

\subsection{Scene Construction}
\label{sec:scene_construction}

For each object in the scene, we build tracks using SAM3~\citep{sam3}, which propagates object tracks forward using visual feature embeddings. We then lift these points to 3D from all views of the objects to obtain a 3D point cloud. Obtaining precise segmentations from our 3FPS video stream at 512x512 resolution poses a challenge (Figure~\ref{fig:seg_comparison}), leading to spurious background points that massively inflate the object's point cloud.

To address this, we implement point cloud filtering in two steps. First, we discard points that are seen in fewer than 50\% of image segmentations (without penalizing points that are missing due to occlusion). Next, we use outlier removal to drop points far away from the central mass of the object. See Appendix~\ref{app:filtering} for details.
% Main results table
\begin{table*}[!t]
\centering
\caption{Main results on all 2,950 questions of \ours{}-Bench across 15 spatial question types. \textbf{PF\%}: parse failure rate; parse failures are counted as wrong. Best per column is \textbf{bold} and second best is \underline{underlined}, computed separately above ($\geq$100B param) and below ($<$100B param) the dividing line. RGB rows receive the video frames with SAM3 overlays; tool-use rows receive no images. \fcolorbox{blue!25}{blue!6}{\strut Closed-source frontier models in blue.} \fcolorbox{gray!30}{gray!15}{\strut Ours in gray.}}
\label{tab:main_results}
\tiny
\setlength{\tabcolsep}{1.5pt}
\resizebox{\textwidth}{!}{%
\begin{tabular}{@{}l|c|c|ccccc!{\color{gray!50}\vrule}c|ccccccc!{\color{gray!50}\vrule}c|ccc!{\color{gray!50}\vrule}c|c@{}}
\toprule
 & & & \multicolumn{6}{c|}{Multiple Choice (Acc $\uparrow$)} & \multicolumn{8}{c|}{Distance (MRA $\uparrow$)} & \multicolumn{4}{c|}{Volume (MRA $\uparrow$)} & \\
\cmidrule(lr){4-9} \cmidrule(lr){10-17} \cmidrule(lr){18-21}
 & \rotatebox{70}{Depth}
 & \rotatebox{70}{PF\%\,$\downarrow$}
 & \rotatebox{70}{Gap Fit}
 & \rotatebox{70}{Nearest}
 & \rotatebox{70}{Top Higher}
 & \rotatebox{70}{Wh.\ Taller}
 & \rotatebox{70}{Wh.\ Longer}
 & \rotatebox{70}{MC Avg}
 & \rotatebox{70}{How Far}
 & \rotatebox{70}{Far From Me}
 & \rotatebox{70}{How Long}
 & \rotatebox{70}{Much Taller}
 & \rotatebox{70}{Much Longer}
 & \rotatebox{70}{Fly Dist.}
 & \rotatebox{70}{Walk Dist.}
 & \rotatebox{70}{Dist Avg}
 & \rotatebox{70}{Volume}
 & \rotatebox{70}{Pour Left.}
 & \rotatebox{70}{Pour Room}
 & \rotatebox{70}{Vol Avg}
 & \rotatebox{70}{Avg} \\
\midrule
\rowcolor{blue!6} GPT-5.5 & \ding{55} & \textbf{0.0} & \textbf{80.3} & 64.0 & \textbf{91.2} & \textbf{94.2} & \underline{87.6} & \textbf{83.5} & 46.4 & 46.9 & \textbf{73.5} & \textbf{61.8} & \textbf{57.5} & 56.8 & 44.0 & \underline{55.3} & 34.0 & \underline{21.2} & \textbf{26.9} & \underline{27.4} & \textbf{59.1} \\
\rowcolor{blue!6} Gemini 2.5 Pro & \ding{55} & 2.8 & \underline{78.7} & 43.0 & 84.8 & 89.8 & 86.7 & 76.6 & 33.5 & 26.0 & 61.5 & 43.8 & 43.2 & 40.8 & 27.0 & 39.4 & 30.6 & 12.6 & 15.2 & 19.5 & 47.8 \\
\rowcolor{blue!6} Gemini 3 Flash & \ding{55} & \underline{0.1} & 72.1 & 49.5 & 82.5 & \underline{91.6} & \textbf{89.8} & \underline{77.1} & 31.7 & 20.6 & \underline{72.3} & \underline{60.0} & \underline{52.8} & 45.0 & 40.4 & 46.1 & \textbf{42.3} & 15.6 & 24.0 & 27.3 & 52.7 \\
\rowcolor{blue!6} Gemini 3.1 Pro & \ding{55} & 1.3 & 70.5 & 53.8 & \underline{88.5} & 86.2 & 84.4 & 76.7 & 27.3 & 31.9 & 65.0 & 48.4 & 44.3 & 42.7 & 38.4 & 42.6 & 26.8 & 9.6 & 18.4 & 18.2 & 49.1 \\
Qwen3-VL 235B-A22B Thinking & \ding{55} & \underline{0.1} & 75.4 & 57.0 & 80.2 & 86.2 & 81.3 & 76.0 & 40.5 & 35.5 & 51.7 & 46.1 & 39.2 & 54.9 & 40.1 & 44.0 & 27.3 & 8.6 & 16.1 & 17.3 & 49.3 \\
CuTR+Tools (Qwen3-VL 235B-A22B Thinking) & \checkmark & 1.9 & 57.4 & \textbf{75.8} & 77.0 & 72.0 & 76.4 & 71.7 & \underline{53.5} & \underline{68.5} & 37.5 & 19.8 & 19.5 & \textbf{62.0} & \underline{45.1} & 43.7 & 35.2 & 14.9 & 16.5 & 22.2 & 48.7 \\
\rowcolor{gray!15} \ours{} (Qwen3-VL 235B-A22B Thinking) & \checkmark & 2.1 & 73.8 & \underline{73.7} & 82.0 & 76.4 & 71.6 & 75.5 & \textbf{55.7} & \textbf{70.5} & 62.8 & 54.5 & 41.4 & \underline{58.6} & \textbf{46.4} & \textbf{55.7} & \underline{40.6} & \textbf{22.8} & \underline{24.8} & \textbf{29.4} & \underline{57.0} \\
\midrule
Qwen3-VL 8B text-only & \ding{55} & \textbf{0.0} & 52.5 & 40.3 & 59.9 & 49.8 & 49.8 & 50.4 & 17.6 & 21.4 & 37.5 & 22.6 & 32.3 & 5.7 & 3.6 & 20.1 & 35.2 & 0.3 & 1.4 & 12.3 & 28.7 \\
Video-3D LLM & \checkmark & 55.9 & 37.7 & 49.5 & 35.0 & 37.8 & 48.4 & 41.7 & 4.8 & 11.8 & 1.2 & 0.4 & 0.0 & 4.4 & 7.7 & 4.3 & 1.6 & 3.5 & 0.0 & 1.7 & 16.3 \\
SpatialRGPT+Median & \checkmark & 20.5 & 45.9 & 22.6 & 50.7 & 35.6 & 32.4 & 37.4 & 18.8 & 18.8 & 28.0 & 10.7 & 9.6 & 0.0 & 0.0 & 12.3 & 24.3 & 0.1 & 1.9 & 8.8 & 20.0 \\
SR-3D & \checkmark & 32.5 & 0.0 & 24.7 & 0.9 & 46.7 & 48.0 & 24.1 & 16.7 & 18.8 & 8.3 & 2.9 & 2.6 & 3.4 & 4.3 & 8.1 & 3.0 & 17.5 & 12.4 & 11.0 & 14.0 \\
CuTR+Tools (Qwen3-VL 8B Instruct) & \checkmark & \underline{6.6} & 62.3 & 60.8 & 61.8 & 60.4 & 67.6 & 62.6 & 55.3 & 52.1 & 34.3 & 21.9 & 21.8 & 35.0 & 20.0 & 34.4 & 31.0 & 16.1 & 16.5 & 21.2 & 41.1 \\
CuTR+Tools (Qwen3.6 35B-A3B) & \checkmark & 7.1 & 60.7 & \underline{74.7} & \textbf{74.2} & \underline{76.4} & \textbf{76.4} & \underline{72.5} & 52.2 & \underline{70.7} & 33.9 & 21.3 & 19.8 & \textbf{56.0} & \textbf{32.4} & 40.9 & 34.6 & 14.9 & 16.5 & 22.0 & \underline{47.7} \\
\rowcolor{gray!15} \ours{} (Qwen3-VL 8B Instruct) & \checkmark & 8.4 & \underline{72.1} & 61.3 & 60.4 & 66.7 & 68.9 & 65.9 & \textbf{58.0} & 53.1 & \underline{58.8} & \underline{48.0} & \underline{37.9} & 32.5 & 16.0 & \underline{43.5} & \underline{36.7} & \underline{19.7} & \underline{24.1} & \underline{26.8} & 47.6 \\
\rowcolor{gray!15} \ours{} (Qwen3.6 35B-A3B) & \checkmark & 7.3 & \textbf{73.8} & \textbf{75.3} & \underline{68.2} & \textbf{80.0} & \underline{73.3} & \textbf{74.1} & \underline{57.1} & \textbf{73.3} & \textbf{62.1} & \textbf{53.2} & \textbf{42.4} & \underline{53.5} & \underline{31.0} & \textbf{53.2} & \textbf{39.9} & \textbf{24.8} & \textbf{25.6} & \textbf{30.1} & \textbf{55.6} \\
\bottomrule
\end{tabular}%
}
\end{table*}

\subsection{Multi-Step Tool Calling}
\label{sec:tool_calling}

After constructing our scene, we perform inference using multi-step tool calling. Our tools utilize computations performed on the scene's 3D object point clouds. Our tools are built on top of (1) a bounding box representation for reasoning over position and object extent, and (2) a mesh-based representation used for volumetric reasoning. Both representations are computed directly from the point cloud.

\textbf{Bounding Box Representation:} For distance-based Q\&A, we compute gravity-aligned bounding boxes over the point clouds. The vertical extent is the vertical range of the point cloud. The horizontal orientation is determined by 2D PCA on the XZ-plane projection of the point cloud.

\textbf{Mesh-Based Representation:} Estimating the object's volume presents a challenge, as many objects are only seen from a few angles (Figure~\ref{fig:dataset_stats}). This makes volumetric estimation from a point cloud challenging, as many objects will have massive holes. To address this, we use SAM3D~\citep{sam3d} to produce an estimated mesh for objects using the first SAM3 segmentation of the object as input. Because SAM3D does not provide meshes in metric space, we need to rescale it. To do this, we scale the mesh in each dimension uniformly by the factor
\begin{equation}
s = \left(\frac{V_{\text{bbox}}}{V_{\text{mesh}}}\right)^{1/3},
\label{eq:mesh_scale}
\end{equation}
where $V_{\text{bbox}}$ is the volume of the object's gravity-aligned bounding box (computed from the point cloud, discussed above) and $V_{\text{mesh}}$ is the volume of the SAM3D mesh's bounding box. We then compute the functional volume of the scaled mesh using voxel-based cavity detection.

Using the above building blocks, \ours{} provides seven composable spatial tools to the LLM as function definitions:
\begin{enumerate}[nosep]
\item \texttt{list\_objects()} --- discover available objects and their IDs.
\item \texttt{get\_distance(id1, id2)} --- Euclidean distance between the closest points on bounding boxes.
\item \texttt{get\_position(id)} --- 3D position of an object's bounding-box center.
\item \texttt{get\_object\_size(id)} --- gravity-aligned bounding box dimensions.
\item \texttt{get\_object\_volume(id)} --- computes the functional volume of the scaled mesh.
\item \texttt{get\_distance\_from\_me(id)} --- distance from the current camera position to the closest point on an object's bounding box.
\item \texttt{get\_my\_position()} --- current camera position in the scene.
\end{enumerate}
Note that the design follows a resolve-first pattern. The LLM must identify objects by numeric ID (via \texttt{list\_objects}) before querying their spatial properties.

%----------------------------------------------------------------------
\section{Experiments}
\label{sec:experiments}

In Section~\ref{sec:main_results}, we compare \ours{} against other methods on \ours{}-Bench, to (1) demonstrate the effectiveness of \ours{}, and (2) demonstrate the challenges presented by \ours{}-Bench. In Section~\ref{sec:error_analysis}, we provide error analysis. In Section~\ref{sec:volumetric}, we discuss the particularly challenging case of volumetric estimation. We show additional analysis of runtime considerations in Appendix~\ref{app:runtime}.

\newcommand{\reconimg}[1]{\includegraphics[width=0.24\textwidth]{figures/recon/#1}}
\newcommand{\reconref}[1]{\includegraphics[width=0.145\textwidth]{figures/recon/#1}}

%----------------------------------------------------------------------

\begin{figure*}[t]
\centering
\setlength{\tabcolsep}{2pt}
\scriptsize
\renewcommand{\arraystretch}{1.2}
\newcommand{\good}[1]{\colorbox{tealgreen!10}{\textcolor{tealgreen!80!black}{\textbf{#1}}}}
\newcommand{\bad}[1]{\colorbox{terra!7}{\textcolor{terra!90!black}{#1}}}
\newcommand{\fail}[1]{\colorbox{gray!10}{\textcolor{gray}{\textit{#1}}}}
\newcommand{\tc}[1]{\texttt{\textcolor{slateblue}{#1}}}
\resizebox{\textwidth}{!}{%
\begin{tabular}{@{}>{\raggedright\arraybackslash}m{2.8cm}|>{\raggedright\arraybackslash}m{3.95cm}|>{\raggedright\arraybackslash}m{3.95cm}|>{\centering\arraybackslash}m{2.6cm}|>{\centering\arraybackslash}m{2.8cm}|>{\raggedright\arraybackslash}m{3.4cm}@{}}
\toprule
\textbf{Question} & \textbf{\ours{} + Qwen3.6 35B-A3B (tools)} & \textbf{CuTR+Tools (Qwen3.6 35B-A3B)} & \textbf{GPT-5.5 (RGB)} & \textbf{Video-3D LLM} & \textbf{SpatialRGPT+Median} \\
\midrule
\textit{How many meters apart are the muffin pan and the wooden fork?} \newline \textbf{GT: 0.206\,m}
&
\roboticon\;\tc{list\_objects()} \newline
\textcolor{gray}{$\to$ 31 objs: 37:muffin pan, 62:wooden fork\,\ldots} \newline
\roboticon\;\tc{get\_distance(id1=37, id2=62)} \newline
\textcolor{gray}{$\to$ error: non-numeric ID} \newline
\roboticon\;\tc{get\_distance(37, 62)} \newline
\textcolor{gray}{$\to$ 0.19\,m} \newline
\good{0.19\,m (7.8\% err)}
&
\roboticon\;\tc{list\_objects()} \newline
\textcolor{gray}{$\to$ 31 objs: 37:muffin pan, 62:wooden fork\,\ldots} \newline
\roboticon\;\tc{get\_distance(id1=37, id2=62)} \newline
\textcolor{gray}{$\to$ error: non-numeric ID} \newline
\roboticon\;\tc{get\_distance(37, 62)} \newline
\textcolor{gray}{$\to$ 0.13\,m} \newline
\bad{0.13\,m (36.9\% err)}
&
\parbox{2.6cm}{\centering RGB, 50 frames\\$\downarrow$\\\roboticon\\$\downarrow$\\\bad{0.35\,m (69.9\% err)}}
&
\parbox{2.6cm}{\centering RGB-D+Pose+GT box\\$\downarrow$\\\roboticon\\$\downarrow$\\\fail{``muffin pan''}\newline{\tiny\textcolor{gray}{(parse fail)}}}
&
RGB-D+Bbox$_0$ $\to$ \roboticon\ $\to$ $y_0$ \newline RGB-D+Bbox$_1$ $\to$ \roboticon\ $\to$ $y_1$ \newline RGB-D+Bbox$_2$ $\to$ \roboticon\ $\to$ $y_2$ \newline RGB-D+Bbox$_3$ $\to$ \roboticon\ $\to$ $y_3$ \newline med$(y_i)\!=$ \bad{1.1\,m (434.0\% err)}
\\
\midrule
\textit{What is the volume of the wooden bowl, in liters?} \newline \textbf{GT: 4.55\,L}
&
\roboticon\;\tc{list\_objects()} \newline
\textcolor{gray}{$\to$ 31 objs: 27:wooden bowl\,\ldots} \newline
\roboticon\;\tc{get\_object\_volume(id=27)} \newline
\textcolor{gray}{$\to$ error: non-numeric ID} \newline
\roboticon\;\tc{get\_object\_volume(27)} \newline
\textcolor{gray}{$\to$ 3.920\,L} \newline
\good{3.92\,L (13.8\% err)}
&
\roboticon\;\tc{list\_objects()} \newline
\textcolor{gray}{$\to$ 31 objs: 27:wooden bowl\,\ldots} \newline
\roboticon\;\tc{get\_object\_volume(id=27)} \newline
\textcolor{gray}{$\to$ error: non-numeric ID} \newline
\roboticon\;\tc{get\_object\_volume(27)} \newline
\textcolor{gray}{$\to$ 5.539\,L} \newline
\good{5.539\,L (21.7\% err)}
&
\parbox{2.6cm}{\centering RGB, 50 frames\\$\downarrow$\\\roboticon\\$\downarrow$\\\bad{3.0\,L (34.1\% err)}}
&
\parbox{2.6cm}{\centering RGB-D+Pose+GT box\\$\downarrow$\\\roboticon\\$\downarrow$\\\fail{``no''}\newline{\tiny\textcolor{gray}{(parse fail)}}}
&
RGB-D+Bbox$_0$ $\to$ \roboticon\ $\to$ $y_0$ \newline RGB-D+Bbox$_1$ $\to$ \roboticon\ $\to$ $y_1$ \newline RGB-D+Bbox$_2$ $\to$ \roboticon\ $\to$ $y_2$ \newline RGB-D+Bbox$_3$ $\to$ \roboticon\ $\to$ $y_3$ \newline med$(y_i)\!=$ \bad{0.5\,L (89.0\% err)}
\\
\midrule
\textit{Between the mocha cake and the white vase, which one is taller?} \newline \textbf{GT: white vase}
&
\roboticon\;\tc{list\_objects()} \newline
\textcolor{gray}{$\to$ 9 objs: 15:mocha cake, 4:white vase\,\ldots} \newline
\roboticon\;\tc{get\_object\_size(id=15)} \newline
\textcolor{gray}{$\to$ error: non-numeric ID} \newline
\roboticon\;\tc{get\_object\_size(15)} \newline
\textcolor{gray}{$\to$ w0.26\,h0.10\,d0.23\,m} \newline
\roboticon\;\tc{get\_object\_size(4)} \newline
\textcolor{gray}{$\to$ w0.10\,h0.16\,d0.08\,m} \newline
\good{white vase \checkmark}
&
\roboticon\;\tc{list\_objects()} \newline
\textcolor{gray}{$\to$ 9 objs: 15:mocha cake, 4:white vase\,\ldots} \newline
\roboticon\;\tc{get\_object\_size(id=15)} \newline
\textcolor{gray}{$\to$ error: non-numeric ID} \newline
\roboticon\;\tc{get\_object\_size(15)} \newline
\textcolor{gray}{$\to$ w0.30\,h0.09\,d0.36\,m} \newline
\roboticon\;\tc{get\_object\_size(4)} \newline
\textcolor{gray}{$\to$ w0.09\,h0.13\,d0.07\,m} \newline
\good{white vase \checkmark}
&
\parbox{2.6cm}{\centering RGB, 50 frames\\$\downarrow$\\\roboticon\\$\downarrow$\\\good{white vase \checkmark}}
&
\parbox{2.6cm}{\centering RGB-D+Pose+GT box\\$\downarrow$\\\roboticon\\$\downarrow$\\\fail{``no''}\newline{\tiny\textcolor{gray}{(parse fail)}}}
&
RGB-D+Bbox$_0$ $\to$ \roboticon\ $\to$ $y_0$ \newline RGB-D+Bbox$_1$ $\to$ \roboticon\ $\to$ $y_1$ \newline RGB-D+Bbox$_2$ $\to$ \roboticon\ $\to$ $y_2$ \newline RGB-D+Bbox$_3$ $\to$ \roboticon\ $\to$ $y_3$ \newline maj$(y_i)\!=$ \bad{mocha cake \ding{55}}
\\
\midrule
\textit{How many meters apart are the grey sofa and the white picture ledge?} \newline \textbf{GT: 5.73\,m} \newline \textcolor{terra}{\tiny never tracked: white picture ledge}
&
\roboticon\;\tc{list\_objects()} \newline
\textcolor{gray}{$\to$ 26 objs: 53:grey sofa, 4:picture ledge\,\ldots} \newline
\roboticon\;\tc{get\_distance(id1=53, id2=4)} \newline
\textcolor{gray}{$\to$ error: non-numeric ID} \newline
\roboticon\;\tc{get\_distance(53, 4)} \newline
\textcolor{gray}{$\to$ 0.99\,m} \newline
\textcolor{terra}{\textit{no white picture ledge in scene; used 4:picture ledge}} \newline
\bad{0.99\,m (82.7\% err)}
&
\roboticon\;\tc{list\_objects()} \newline
\textcolor{gray}{$\to$ 26 objs: 53:grey sofa, 4:picture ledge\,\ldots} \newline
\roboticon\;\tc{get\_distance(id1=53, id2=4)} \newline
\textcolor{gray}{$\to$ error: non-numeric ID} \newline
\roboticon\;\tc{get\_distance(53, 4)} \newline
\textcolor{gray}{$\to$ 0.96\,m} \newline
\textcolor{terra}{\textit{no white picture ledge in scene; used 4:picture ledge}} \newline
\bad{0.96\,m (83.3\% err)}
&
\parbox{2.6cm}{\centering RGB, 50 frames\\$\downarrow$\\\roboticon\\$\downarrow$\\\good{6.4\,m (11.7\% err)}}
&
\parbox{2.6cm}{\centering RGB-D+Pose+GT box\\$\downarrow$\\\roboticon\\$\downarrow$\\\bad{2.46\,m (57.1\% err)}}
&
RGB-D+Bbox$_0$ $\to$ \roboticon\ $\to$ $y_0$ \newline RGB-D+Bbox$_1$ $\to$ \roboticon\ $\to$ $y_1$ \newline RGB-D+Bbox$_2$ $\to$ \roboticon\ $\to$ $y_2$ \newline RGB-D+Bbox$_3$ $\to$ \roboticon\ $\to$ $y_3$ \newline med$(y_i)\!=$ \bad{1.5\,m (73.8\% err)}
\\
\bottomrule
\end{tabular}%
}
\caption{Qualitative examples. Rows 1--3: one question of each category. Row 4: a distance question whose white picture ledge SAM3 never tracked, so it is missing from the scene; both tool-use agents substitute a different object (a picture ledge) and answer incorrectly, while GPT-5.5 sees the ledge in the frames. The robot icon denotes an LLM call. \ours{} and CuTR answer via multi-step tool calling (verbatim traces, malformed calls included); GPT-5.5 answers from RGB frames; Video-3D LLM is an end-to-end model given ground-truth boxes; SpatialRGPT+Median runs per-frame inference and aggregates by median/majority vote. \textcolor{tealgreen}{\textbf{Teal}}: correct or ${\leq}$25\% error. \textcolor{terra}{Red}: incorrect or ${>}$25\% error. \textcolor{gray}{Gray}: parse failure.}
\label{fig:qualitative}
\end{figure*}

\subsection{Main Results}
\label{sec:main_results}

Our main results appear in Table~\ref{tab:main_results}. We experiment with using R3D with a Qwen~\citep{qwen3vl,qwen3,qwen36} model at three different sizes. For efficiency, we evaluate R3D without image inputs, relying on tool use for the visual signal. In ablations, we found no difference in accuracy with and without image inputs on R3D-Bench. Thus our method is compatible with image inputs, but we omit them in R3D evaluations for computational efficiency.

We divide our results into two sections, for models with $\geq$100B parameters and $<$100B parameters. For multiple-choice questions, we report accuracy. For numerical questions, we report Mean Relative Accuracy (MRA)~\citep{vsibench}, which averages accuracy over 10 thresholds:
\begin{equation}
\textstyle\text{MRA}(\hat{y}, y) = \tfrac{1}{10}\sum_{k=0}^{9} \mathbf{1}\!\bigl[\tfrac{|\hat{y}-y|}{|y|} < 1 {-} (0.50 {+} 0.05k)\bigr].
\label{eq:mra}
\end{equation}
\textbf{Comparison with Large State-of-the-Art LLMs:} We compare \ours{} with large closed-source models (Gemini~\citep{gemini25,gemini3flash,gemini31pro}, GPT~\citep{gpt55}), and an open-source Qwen~\citep{qwen3vl} model in Table~\ref{tab:main_results}. These RGB models do not have a provision for accepting depth, so we do not provide it. For fair comparison, we overlay RGB images with our SAM3 masks and bounding box labels for the queried objects, to disambiguate the object of interest.

With Qwen3-VL 235B-A22B Thinking, \ours{} outperforms Gemini and approaches the performance of the strongest large closed-source model, GPT-5.5, within 2.1\% (absolute). Our volumetric estimation method gives the best volume average of all methods, demonstrating the value of using SAM3D meshes.

We analyze the breakdown of when our model succeeds and fails in Appendix~\ref{app:additional_evaluations}. We find that, on examples with object queries that demonstrate good performance from the tracker, our method outperforms all baselines (including GPT-5.5). On examples where tracker performance lags, we fall behind the RGB baselines, which can still see objects that SAM3 fails to track.

\textbf{Comparison with Alternative 3D Methods:} We compare with other depth-based methods in Table~\ref{tab:main_results}. We compare with Video-3D LLM, which finetunes a multimodal LLM on RGB-D and pose videos for spatial understanding tasks. We evaluate it with ground truth ADT scene boxes to rule out poor box quality explaining its poor performance (see Appendix~\ref{app:baselines} for details). Unfortunately, we find that this model exhibits parse failures on 55.9\% of examples (see Figure~\ref{fig:qualitative}), likely because the model was primarily trained on qualitative (not quantitative) relationships. On the questions that don't exhibit parse failures, it still frequently underperforms the Qwen3-VL 8B text-only baseline. This underscores the need for more quantitative reasoning datasets for evaluating 3D methods.

We additionally evaluate the multi-view region-prompted model SR-3D~\citep{sr3d}. We remap region names that it outputs, such as ``Region [0]'', to object names, to reduce parse errors. However, it still refuses to answer questions such as ``gap fit'', replying with object names or descriptions rather than yes/no answers, yielding only parse failures for these questions.

We also construct a strong ``CuTR+Tools'' baseline, in which CuTR~\citep{cubifyanything} acts as the perception engine in place of our depth-lifted point cloud and SAM3D representation, with the tool interface held fixed.\footnote{We sought to compare directly with MM-Spatial~\citep{mmspatial_model}, which performs multi-view spatial Q\&A, but at the time of writing, its code release omits model weights. Therefore, we compare against CuTR, which MM-Spatial builds upon. See Appendix~\ref{app:baselines} for more details.} We find that our method outperforms ``CuTR+Tools'' at every model size. The biggest performance gap appears on object length (How Long, Much Taller, Much Longer). Visualization inspection suggests that CuTR typically predicts boxes that are too small compared to ground truth.

We additionally adapt SpatialRGPT~\citep{spatialrgpt}, an image-based method, to video by aggregating its per-frame predictions. Performance is particularly low on questions that require seeing multiple objects (Much Taller, Much Longer, Fly Dist., Walk Dist.) as the model may not see these objects simultaneously. See Appendix~\ref{app:baselines} for additional details on all baselines.

\textbf{R3D-Bench is Challenging:} Overall performance peaks at 59.1\%, even when using MRA as the primary metric (which gives partial credit for measurement errors up to 50\%, see Equation~\ref{eq:mra}). Volumetric reasoning peaks at 30.1\%, indicating the particular challenge of these questions and opportunities for algorithmic improvement. Text-only Qwen3-VL 8B can achieve 35.2\% on volume estimation by guessing reasonable values, but achieves less than 2\% on the two pouring question types.

\begin{table*}[t]
\centering
\caption{Error analysis of \ours{} for three model sizes (235B = Qwen3-VL 235B-A22B Thinking, 35B = Qwen3.6 35B-A3B, 8B = Qwen3-VL 8B Instruct), on all 2,950 questions. N is the number of incorrect answers in each category and \% is the share of that model's incorrect answers. Columns follow the order in which the labels are assigned (Section~\ref{sec:error_analysis}).}
\label{tab:error_analysis}
\scriptsize
\setlength{\tabcolsep}{3pt}
\begin{tabular}{@{}lr|rr|rr|rr|rr|rr@{}}
\toprule
 & & \multicolumn{2}{c|}{Obj.\ missing} & \multicolumn{2}{c|}{Other parse} & \multicolumn{2}{c|}{Tool calling} & \multicolumn{2}{c|}{Measurement} & \multicolumn{2}{c}{Reasoning} \\
\cmidrule(lr){3-4} \cmidrule(lr){5-6} \cmidrule(lr){7-8} \cmidrule(lr){9-10} \cmidrule(lr){11-12}
Model & Wrong & N & \% & N & \% & N & \% & N & \% & N & \% \\
\midrule
\ours{} (235B) & 1,233 & 902 & 73.2 & 0 & 0.0 & 54 & 4.4 & 246 & 20.0 & 31 & 2.5 \\
\ours{} (35B) & 1,279 & 924 & 72.2 & 3 & 0.2 & 76 & 5.9 & 241 & 18.8 & 35 & 2.7 \\
\ours{} (8B) & 1,552 & 1,034 & 66.6 & 14 & 0.9 & 118 & 7.6 & 228 & 14.7 & 158 & 10.2 \\
\bottomrule
\end{tabular}
\end{table*}

\begin{figure*}[t]
\centering
\setlength{\tabcolsep}{3pt}
\scriptsize
\def\reconimg#1{\includegraphics[width=0.24\textwidth]{\ntbAssetDir/recon/#1}}
\def\reconref#1{\includegraphics[width=0.145\textwidth]{\ntbAssetDir/recon/#1}}
\begin{tabular}{@{}>{\centering\arraybackslash}m{0.8cm}>{\centering\arraybackslash}m{0.145\textwidth}>{\centering\arraybackslash}m{0.24\textwidth}>{\centering\arraybackslash}m{0.24\textwidth}>{\centering\arraybackslash}m{0.24\textwidth}@{}}
\toprule
& \textbf{SAM3 Seg} & \textbf{R3D Pointcloud} & \textbf{R3D Mesh} & \textbf{Ground Truth Mesh} \\
\midrule
\rotatebox{90}{\scriptsize\shortstack{large coffee canister\\clean\_seq150}} &
\reconref{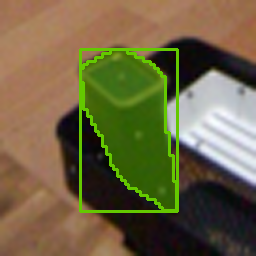} &
\reconimg{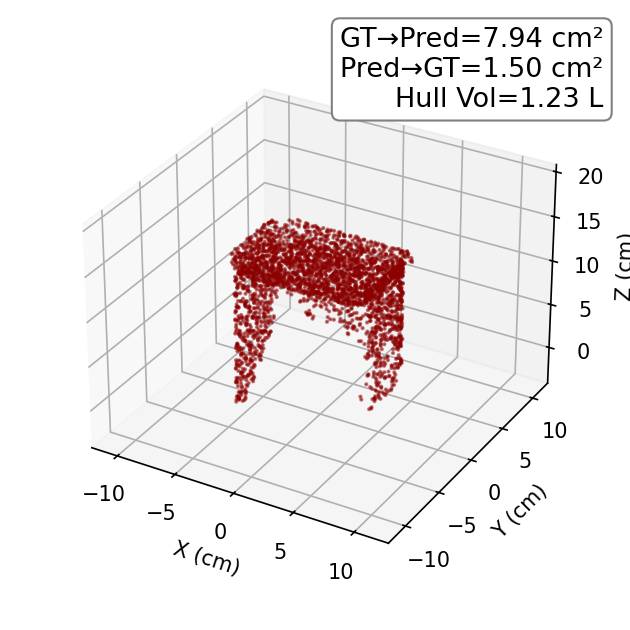} &
\reconimg{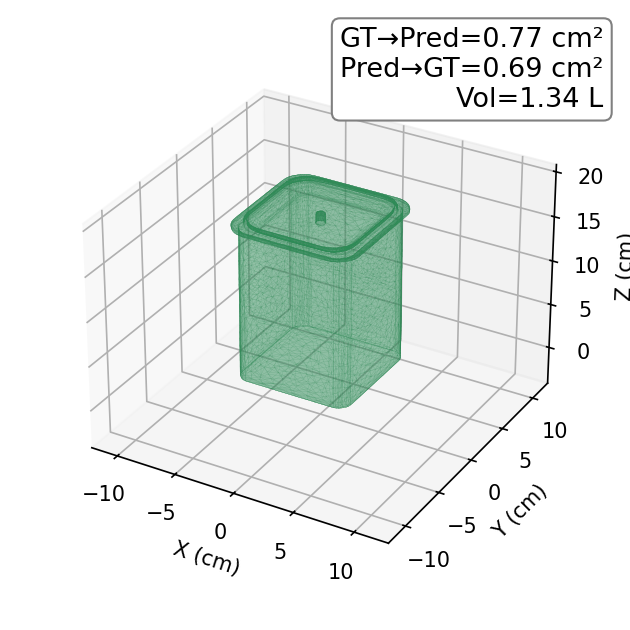} &
\reconimg{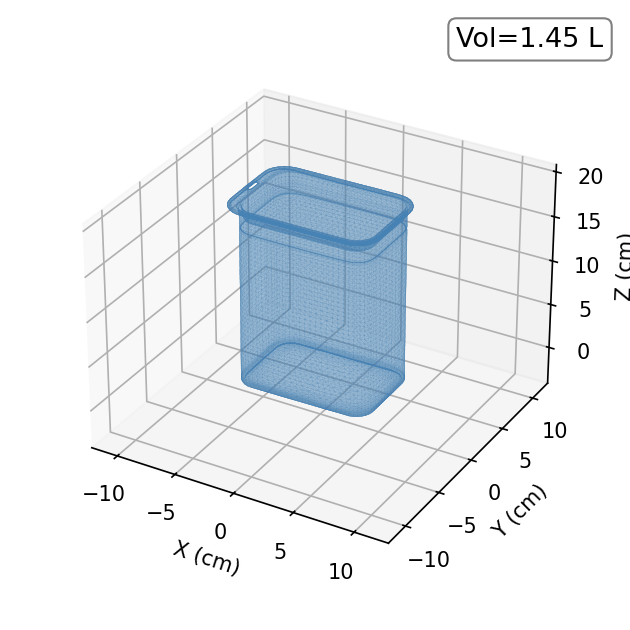} \\
\rotatebox{90}{\scriptsize\shortstack{wooden bowl\\clean\_seq148}} &
\reconref{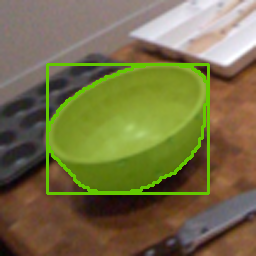} &
\reconimg{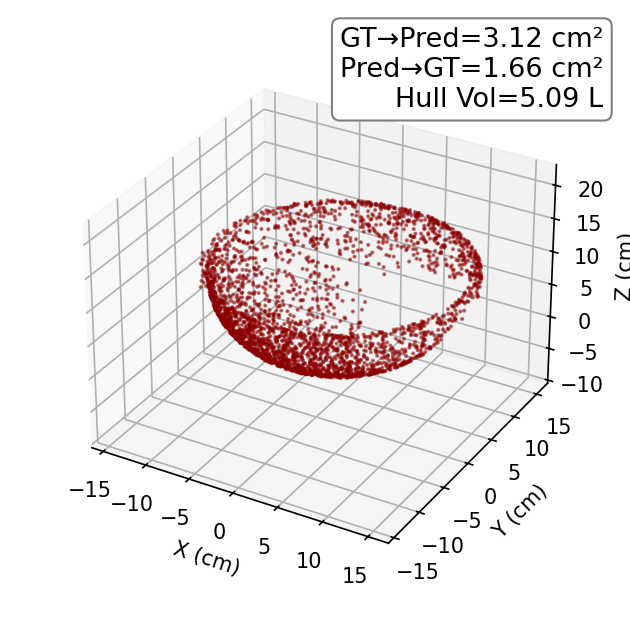} &
\reconimg{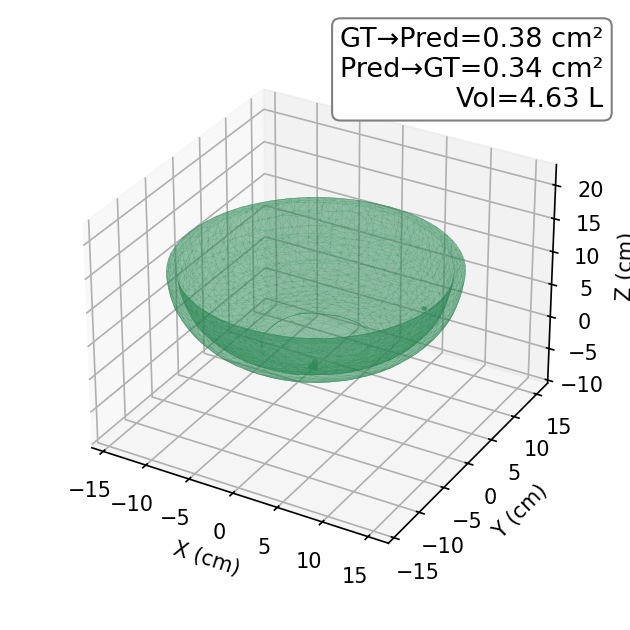} &
\reconimg{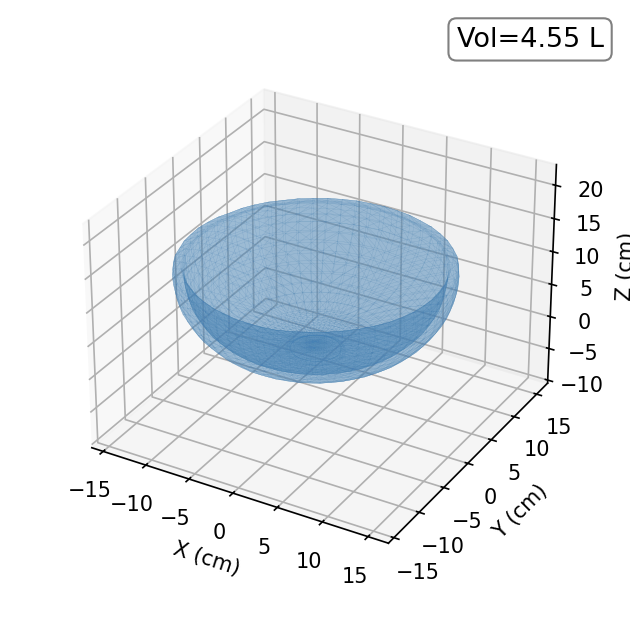} \\
\bottomrule
\end{tabular}
\caption{3D reconstruction comparison. SAM3 Seg: the SAM3 segmentation crop passed to SAM3D. Pointcloud: the depth-lifted object points of the scene (fused, multi-view vote, KNN). R3D Mesh: SAM3D mesh rescaled by the cube root of the scene-box to mesh-box volume ratio. GT: ground-truth mesh from ADT. We show Chamfer distance, volumes, and the convex hull volume of the point cloud.}
\label{fig:recon_comparison}
\end{figure*}

\subsection{Error Analysis}
\label{sec:error_analysis}
We analyze the errors of \ours{} in Table~\ref{tab:error_analysis}. An answer is wrong if it cannot be parsed, a numeric answer is off by more than 25\%, or a multiple-choice answer is incorrect. Each wrong answer receives exactly one label, from the first of the following rules that applies:
\begin{enumerate}[nosep]
  \item \textbf{Object missing:} an object referenced by the question is absent from the reconstructed scene because SAM3 never tracked it, so no tool can measure it. The model either gives a parse failure or guesses.
  \item \textbf{Other parse:} no answer can be extracted from the response, and every referenced object is in the scene.
  \item \textbf{Tool calling:} the model never obtained a quantity the question requires (e.g.\ it answers a volume question without calling \texttt{get\_object\_volume}).
  \item \textbf{Measurement:} the model obtained the required quantities, but they are too inaccurate to answer correctly. We run a fixed program for the question type (e.g.\ subtracting two heights) on the tool outputs in the model's trace; if the program's answer is also wrong, the error is a measurement error.
  \item \textbf{Reasoning:} the program answers correctly from the model's own tool outputs, but the model answers incorrectly, e.g.\ because of an arithmetic mistake, a wrong comparison, or querying the wrong object.
\end{enumerate}
Tool calling and reasoning errors differ in whether the needed information reached the model. With a tool calling error it never did, while with a reasoning error the model had sufficiently accurate information and still answered incorrectly.

Missing objects cause most errors (66.6--73.2\%), followed by measurement (14.7--20.0\%). All models share the same perception system, so better tracking and reconstruction would reduce errors for all of them. Qwen3-VL 8B is more limited by reasoning than the larger models, with 158 reasoning errors versus 35 (35B) and 31 (235B).

Additionally, we analyze the scene reconstruction quality compared to ground truth (Appendix~\ref{app:scene_quality}), the impact of scene construction (Appendix~\ref{app:scene_ablation}) and of the tool interface (Appendix~\ref{app:tool_ablation}) on Q\&A performance, and the upper bounds given ground-truth scenes and tools (Appendix~\ref{app:gt_evals}).

\subsection{Volumetric Reasoning Performance}
\label{sec:volumetric}
Volumetric questions present the greatest challenge and the lowest accuracy numbers for nearly all models (Table~\ref{tab:main_results}) due to the sparse viewpoints present for objects. In Figure~\ref{fig:recon_comparison}, we show two examples of the inputs used to create our 3D representation: (1) the SAM3 mask used by SAM3D, and (2) the point cloud derived from multiple views used for scaling via Equation~\ref{eq:mesh_scale}. We also show the mesh produced by R3D (using SAM3D and scaling), compared to the ground truth. SAM3D helps improve the representation by filling in missing details, but still struggles to precisely reconstruct objects from the single-view segmentation alone. Additional analysis measures the reconstruction quality against the ground truth meshes using Chamfer distance in
Appendix~\ref{app:recon}. We also compare our SAM3D approach to using bounding boxes for volume estimation for both \ours{} and CuTR in Appendix~\ref{app:baselines}.

%----------------------------------------------------------------------
\section{Limitations}
\label{sec:limitations}

\ours{}-Bench contains limited variety of scenes, as it was built on top of ADT which lacks scene diversity. \ours{} requires calibrated depth maps and 6-DoF camera pose as input, which are currently available only on devices with active depth sensors and onboard SLAM (e.g.\ augmented reality or mixed reality headsets). This is by design: our focus is on developing evaluations and methods for next-generation wearable devices where calibrated depth and pose are available. Its performance also depends on performance of the underlying tracker.

%----------------------------------------------------------------------
\section{Conclusion}
\label{sec:conclusion}

We present \textbf{\ours{}-Bench}, a benchmark for quantitative spatial reasoning from egocentric video. Our benchmark provides (1) natural egocentric video, (2) calibrated depth and camera pose, and (3) challenging quantitative 3D spatial reasoning Q\&A to provide a realistic benchmark for depth-enabled wearables. \ours{}-Bench contains 2,950 3D quantitative spatial reasoning questions across 15 types over 57 egocentric video sequences. Our dataset provides challenges such as small objects, sparse viewpoints, low resolution, and tracking failures.

We also present \textbf{\ours{}}, a tool calling framework that can be integrated into any tool-enabled LLM without requiring training. On \ours{}-Bench, \ours{} with Qwen3.6 35B-A3B outperforms all baselines $<$100B parameters, achieving 55.6\% mean relative accuracy, a substantial increase over the second-best approach (CuTR+Tools, 47.7\%). In the $\geq$100B parameter domain, our \ours{} Qwen3-VL 235B-A22B Thinking variant improves upon its base model by 7.7\% absolute, achieving 57.0\% accuracy and outperforming all open- and closed-source models except closed-source GPT-5.5 (59.1\%), with which it remains competitive.

%----------------------------------------------------------------------

{\small
\bibliographystyle{assets/plainnat}
\bibliography{references}
}

% Appendix in a single column, so its wide tables keep their single-column layout.
\onecolumn
\appendix

%----------------------------------------------------------------------
\section{Computing Ground-Truth Volume}
\label{app:gt_volume}
\begin{figure}[h]
\centering
\includegraphics[width=\textwidth]{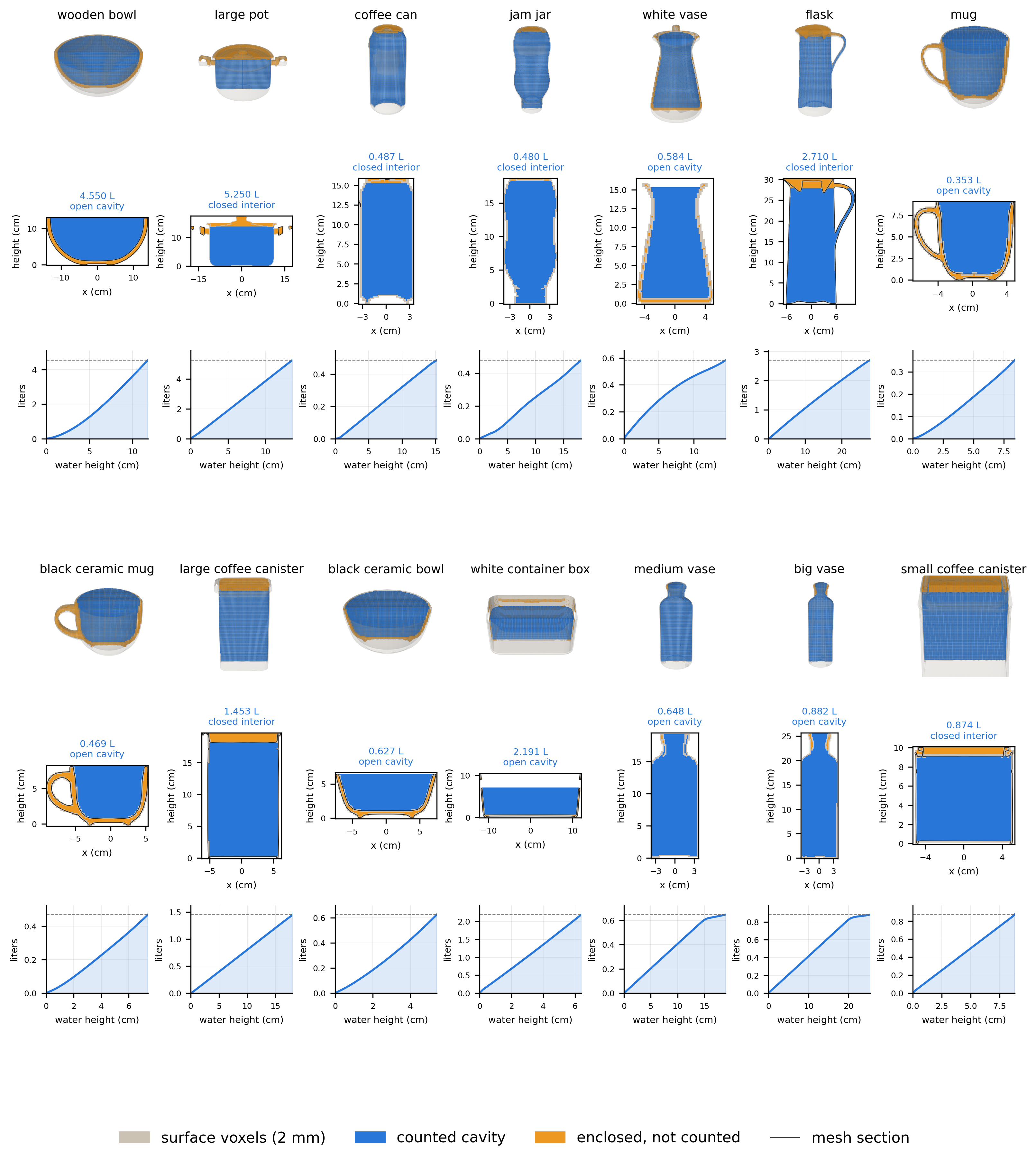}
\caption{Ground-truth functional volume on the 14 container meshes. Top: mesh with the counted cavity (blue) and enclosed regions that are not counted (orange). Middle: vertical section through the voxel grid, with the capacity and whether the cavity is an open cavity or a closed interior. Bottom: counted volume as a function of water height, rising to the capacity (dashed).}
\label{fig:functional_volume}
\end{figure}

Ground-truth capacities come from the ADT object meshes. To compute them, we detect the largest open and the largest closed cavities, and take the larger of the two (to account for both lidded and lid-free objects). In detail, given a mesh, we compute the functional volume as follows (Figure~\ref{fig:functional_volume}):
\begin{enumerate}[nosep]
  \item \textbf{Voxelize:} Voxelize the mesh surface at a 2\,mm pitch, padded by one empty voxel. A voxel is occupied if the mesh surface passes through it; the interior is not filled. Two empty voxels are connected only if they share a face, so water cannot leak diagonally between occupied voxels, and a surface that is one voxel thick is watertight. Holes in the mesh narrower than one voxel are therefore closed.
  \item \textbf{Open Cavity Candidate:} To compute the open cavity candidate, iterate over the voxel layers from the bottom up. An empty voxel at height $k$ holds water if its connected component within heights $\le k$ touches neither the sides nor the bottom of the grid. This way, water fills up to the lowest height at which it can reach the outside, whether over the rim or through a hole, without explicit rim or spout detection. The open candidate is the largest connected component of retained water.
  \item \textbf{Closed Cavity Candidate:} Next, compute the closed cavity candidate. Empty voxels whose component cannot reach any face of the grid are sealed. The closed candidate is the largest sealed component that is more than 6 voxels deep (thinner pockets are likely wall or handle material). Head-space above the highest horizontal slice whose area is at least half of the component's largest slice (e.g.\ a lid dome) is removed. Sealed space is never counted as water in an open cavity, so the sealed double walls and handle tubes of open containers are excluded.
  \item \textbf{Capacity:} The larger of the two candidates is the cavity. Its volume is the number of cavity voxels plus half of the surface voxels face-adjacent to it (correcting the roughly one-voxel under-read per wall of a surface grid), multiplied by the voxel volume.
\end{enumerate}
We visualize the resulting volumetric cavities in Figure~\ref{fig:functional_volume} on all 14 objects for which volume is computed. These objects are chosen as objects that semantically make sense to carry liquid, from among the objects in ADT. In an ablation we find that the volume is not materially affected by changing to 1, 2 and 4\,mm pitch. Volumes change by a median of $+1.5\%$ at 1\,mm and $-3.0\%$ at 4\,mm.

\emph{At evaluation time, we use the same volume computation on SAM3D meshes zero-shot, with no modification of any kind.}

%----------------------------------------------------------------------
\section{Object and Question Filtering}
\label{app:qa_filtering}

We filter objects within each sequence according to the following criteria (Table~\ref{tab:object_funnel}):

\begin{itemize}[nosep]
  \item \textbf{Annotated:} the object has a name and a ground-truth 3D box.
  \item \textbf{Minimum visibility:} in at least 5 sampled frames of the sequence, the object occupies at least 6$^\circ$ of the camera's field of view, is at most 4\,m from the camera, and is at least 50\% visible. (ADT includes visibility annotations.)
  \item \textbf{Unique names:} object names within each sequence must be unique, so a question cannot refer ambiguously to two objects.
  \item \textbf{Static:} the object's box center moves less than 10\,cm over the sequence, so every question has a single well-defined answer.
\end{itemize}

Questions are then generated from these objects. Note that for volumetric questions, we restrict to objects that make sense carrying liquid (see Appendix~\ref{app:gt_volume}). The question generation and filtering are as follows (Table~\ref{tab:question_funnel}):

\begin{itemize}[nosep]
  \item \textbf{Generation cap:} generate at most 100 candidate questions per type per sequence.
  \item \textbf{Yes/no balancing:} for Gap Fit and Top Higher, equal numbers of ``yes'' and ``no'' answers are kept.
  \item \textbf{Obviousness balancing:} for Which Taller and Which Longer, we answer each question with Qwen3-VL 8B (text-only) and keep equal numbers of questions it answers correctly and incorrectly, to avoid a plurality of questions that can be answered from object names alone.
  \item \textbf{Per-sequence cap:} at most 8 questions per type per sequence.
  \item \textbf{Tilted objects:} we remove questions whose object is tilted more than 10$^\circ$ from gravity (so that empty space in the bounding box does not make the top of the object incorrect).
  \item \textbf{Small pour differences:} we remove pour questions whose two capacities differ by less than 0.05\,L.
  \item \textbf{Stratified sample:} a seeded sample by question type gives the final 2,950 questions (Table~\ref{tab:question_types}).
\end{itemize}

\begin{table}[h]
\centering
\small
\caption{Object-stage filters, applied at generation before any question exists. All 57 sequences are recorded in the same ADT apartment, so we count object--sequence pairs and distinct physical objects separately.}
\label{tab:object_funnel}
\resizebox{\textwidth}{!}{%
\begin{tabular}{@{}lrr@{}}
\toprule
Stage & Object--sequence pairs & Distinct objects \\
\midrule
ADT objects & 20,280 & 366 \\
Annotated (name and 3D box) & 19,368 & 350 \\
Minimum visibility ($\geq$5 frames with angular size $\geq$6$^\circ$, distance $\leq$4\,m, visibility $\geq$0.5) & 3,538 & 175 \\
Unique names & 3,131 & 147 \\
Static (box center moves $<$10\,cm) & 2,827 & 147 \\
\midrule
Referenced by the final 2,950 questions & 2,092 & 114 \\
\bottomrule
\end{tabular}%
}
\end{table}

\begin{table}[h]
\centering
\small
\caption{Question funnel from generation to the final 2,950 questions. ``Quality'' stages remove questions with ill-defined answers; ``subsampling'' stages balance or cap the pool.}
\label{tab:question_funnel}
\begin{tabular}{@{}p{0.66\textwidth}lr@{}}
\toprule
Stage & Kind & Questions \\
\midrule
Generation cap ($\leq$100 per type per sequence) & generation & 62,662 \\
Yes/no balancing (Gap Fit, Top Higher) & subsampling & 56,866 \\
Obviousness balancing (Which Taller, Which Longer) & subsampling & 50,116 \\
Per-sequence cap ($\leq$8 per type per sequence) & subsampling & 6,146 \\
Tilted objects ($>$10$^\circ$ from gravity) & quality & 4,811 \\
Small pour differences (capacities differ by $<$0.05\,L) & quality & 4,683 \\
Stratified sample (by question type) & subsampling & 2,950 \\
\bottomrule
\end{tabular}
\end{table}

%----------------------------------------------------------------------
\section{Point Cloud Filtering}
\label{app:filtering}

Here we give the details of the two filtering steps summarized in Section~\ref{sec:scene_construction}. First, we perform voting. For each point $\mathbf{p}_i$, we count the fraction of frames whose segmentation mask contains $\mathbf{p}_i$ when projected into the frame. We discard points seen in fewer than 50\% of segmentations:
\begin{equation}
\mathcal{P}_{\text{vote}} = \left\{\mathbf{p}_i \;\middle|\; \frac{1}{|\mathcal{F}|}\sum_{f \in \mathcal{F}} \mathbf{1}[\mathbf{p}_i \in M_f] \geq 0.5 \right\},
\end{equation}
where $\mathcal{F}$ is the set of frames and $M_f$ is the segmentation mask for frame $f$. Next, we perform KNN outlier removal. For each point in $\mathcal{P}_{\text{vote}}$, we compute $\bar{d}_i$, its mean Euclidean distance to its $k{=}6$ nearest neighbors. We then retain points with
\begin{equation}
\mathcal{P}_{\text{final}} = \left\{\mathbf{p}_i \in \mathcal{P}_{\text{vote}} \;\middle|\; \bar{d}_i \leq 3\,\tilde{d} \right\},
\end{equation}
where $\tilde{d}$ is the median of $\{\bar{d}_i\}$ over all points. This removes isolated outliers from repeated oversegmentations while preserving the main object surface.

%----------------------------------------------------------------------
\section{Inference Time}
\label{app:runtime}

\begin{table}[!ht]
\centering
\caption{Per-question runtime comparison, averaged over all 2,950 questions of \ours{}-Bench. Benchmarked on a single H100 GPU and an AMD EPYC 9004 CPU. SAM3 and depth lifting scale with the number of referenced objects per question (1.875; 29.0\,s and 0.6\,s per object). SAM3D runs on the volume and pour objects that have a mesh (18.9\,s per object); objects SAM3 never tracked have no mask and no mesh (counting every volume and pour object instead gives 4.8\,s). CuTR (22.0\,s) is a per-sequence cost. Model latencies were measured with Qwen3-VL 8B Instruct and Qwen3.6 35B-A3B.}
\label{tab:runtime}
\small
\setlength{\tabcolsep}{6pt}
\begin{tabular}{@{}l|rrrr|r|r@{}}
\toprule
 & \multicolumn{4}{c|}{Avg Preproc.\ Time (s)} & \multicolumn{1}{c|}{Model} & \multicolumn{1}{c}{Total} \\
\cmidrule(lr){2-5}
 & SAM3 & CuTR & Lift & SAM3D & \multicolumn{1}{c|}{Latency (s)} & \multicolumn{1}{c}{(s)} \\

\midrule
Video-3D LLM & 54.4 & 22.0 & --- & --- & 0.8 & 77.2 \\
SpatialRGPT & 54.4 & --- & --- & --- & 3.1 & 57.5 \\
CuTR+Tools (8B) & 54.4 & 22.0 & --- & --- & 2.0 & 78.4 \\
CuTR+Tools (35B) & 54.4 & 22.0 & --- & --- & 5.7 & 82.1 \\
\ours{} (8B) & 54.4 & --- & 1.1 & 3.0 & 2.0 & 60.5 \\
\ours{} (35B) & 54.4 & --- & 1.1 & 3.0 & 5.7 & 64.2 \\
\bottomrule
\end{tabular}
\end{table}

We analyze the runtime performance of our method and baselines in Table~\ref{tab:runtime}. Benchmarking is done on a single H100 GPU and an AMD EPYC 9004 CPU. The average processing times for each query from \ours{}-Bench are shown. All methods rely on SAM3 to provide object tracks. Video-3D LLM and CuTR+Tools additionally use 3D bounding box candidates (which are then matched with SAM3 object identities). For fair comparison, we include the CuTR box generation time as part of Video-3D LLM inference time even though our evaluations in Table~\ref{tab:main_results} used ground truth 3D ADT boxes in an effort to improve accuracy. Each system component is benchmarked separately, resulting in identical values in all columns except Model Latency.

The preprocessing times shown represent latency that can be hidden by running these systems on the scene's objects before the user's query arrives. Model latency represents actual perceived user latency. We find that preprocessing times are dominated by SAM3. Our method can achieve a competitive model latency with Video-3D LLM and SpatialRGPT by using a similar sized model (Qwen3-VL 8B): it is lower than SpatialRGPT's, but higher than Video-3D LLM's due to multiple rounds of tool calling rather than one-shot Q\&A. For our larger and more accurate variants, accuracy increases at the expense of latency.

%----------------------------------------------------------------------
\section{Baseline Details}
\label{app:baselines}

Here we add additional details about our baselines.

\textbf{Video-3D LLM.} We use the spatial grounding inference protocol described
in \citet{video3dllm}, which accepts scene bounding boxes. To rule out box
quality as a source of the observed low performance, we provide ground truth ADT
bounding boxes as the scene bounding boxes.

\textbf{CuTR+Tools.} We attempted to compare with MM-Spatial~\citep{mmspatial_model},
which uses RGB-D and pose with multiview images as input for spatial question
answering. As of the time of experimenting, their code release does not include
model weights. We therefore compare with CuTR~\citep{cubifyanything}, which MM-Spatial is built on top
of, and construct our ``CuTR+Tools'' baseline by combining CuTR with our tool
calling framework. In this setup, CuTR acts as the perception engine, replacing
our depth-lifted point cloud and SAM3D representation. We run CuTR on every
video frame at 3FPS, choosing the highest-confidence bounding box that overlaps
with a SAM3 segmentation as the object representation. This box is used to
compute length, height, position, and volume. This experiment provides a direct
comparison of our \ours{} perception system with a state-of-the-art depth-based
detection system, while keeping the tool interface fixed.

\begin{table}[h]
\centering
\small
\setlength{\tabcolsep}{4pt}
\caption{CuTR and \ours{} with box volumes and with SAM3D volumes (Qwen3.6 35B-A3B, same tools, no images, all 2,950 questions, parse failures counted as wrong). Only \texttt{get\_object\_volume} differs within each method. The last column is the median relative error of the estimated volume (the box volume for Box rows, the rescaled SAM3D functional volume for SAM3D rows) against the ground-truth capacity, over the 163 objects of the volume and pour questions that SAM3 tracks (of 265 referenced; the other 102 are never tracked).}
\label{tab:cutr_sam3d}
\resizebox{\textwidth}{!}{%
\begin{tabular}{@{}llrrrrrr@{}}
\toprule
Scene & Volume from & Volume & Pour Left. & Pour Room & Vol avg & Overall & Volume err. $\downarrow$ \\
\midrule
CuTR & Box & 34.6 & 14.9 & 16.5 & 22.0 & 47.7 & 33.4\% \\
CuTR & SAM3D rescaled to CuTR box & 13.8 & 7.4 & 8.5 & 9.9 & 45.2 & 58.1\% \\
\midrule
\ours{} & Box & 14.0 & 6.5 & 6.1 & 8.8 & 51.3 & 114.1\% \\
\ours{} & SAM3D rescaled to \ours{} box & 39.9 & 24.8 & 25.6 & 30.1 & 55.6 & \textbf{23.2\%} \\
\bottomrule
\end{tabular}%
}
\end{table}

We note that CuTR underperforms when using SAM3D instead of its native boxes (Table~\ref{tab:cutr_sam3d}). We find that CuTR tends to predict boxes that are too small. This balances out the overestimation caused by the bounding box including air outside the object, but causes a problem when this overestimation factor is removed. Per object, CuTR's box volume has a median error of 33.4\% against the ground-truth capacity, while the SAM3D mesh rescaled to CuTR's box has 58.1\%; for \ours{} the order reverses (box 114.1\%, rescaled SAM3D 23.2\%). SAM3D volumes help only when the box they are rescaled to is accurate.

\textbf{SpatialRGPT.} Because few works study reasoning on RGB-D video, we adapt
this image-based spatial understanding work to operate on videos. We run
SpatialRGPT~\citep{spatialrgpt} on every frame, then choose the median (or mode
for multiple-choice questions) of the output distribution as the response. For
fair comparison, we feed SpatialRGPT ground truth depth maps instead of using
monocular depth estimation.

\textbf{SR-3D.} We run the released multi-view SR-3D model~\citep{sr3d}, trained on ScanNet-based data, on up to 32 frames per question, always including the pointing frames. Each frame comes with a point map computed from the ground-truth depth and camera pose. SR-3D normalizes the point maps of each sample to $[-1, 1]$, so its inputs carry no metric scale. The referenced objects are given as region prompts: each object's SAM3 mask in each frame, converted to a filled bounding box. Because SR-3D ignores the requested answer format, we map its free-form answers to our format (e.g.\ ``Region [0] is taller'' becomes the object's name); 32.5\% of its answers still cannot be parsed.

%----------------------------------------------------------------------
\section{Parse Failure and Tracker Performance Ablations}
\label{app:additional_evaluations}
Our main results (Table~\ref{tab:main_results}) count parse failures as wrong. Table~\ref{tab:main_results_all_std} gives the same results with parse failures excluded from scoring, to highlight how they affect results.

\textbf{Impact of Tracking on Performance:} All methods rely on SAM3 to locate the referenced objects. To measure how much of each method's error comes from tracking, we split the 2,950 questions by SAM3 tracking quality. For each referenced object, we compare its SAM3 track to the ground-truth masks over the sequence. The object is considered successfully tracked if its mean IoU is above 0.5 and fewer than 20\% of its tracked frames have IoU below 0.05 with the ground truth (i.e.\ SAM3 is not following a different object). A question is in the \emph{tracker-pass} subset if every object it references meets these criteria (1,153 questions), and in the \emph{tracker-failure} subset otherwise (1,797 questions). Table~\ref{tab:main_results_tracker_pass} reports results on the tracker-pass questions and Table~\ref{tab:main_results_tracker_failure} on the tracker-failure questions.

\begin{table}[!t]
\centering
\caption{Main results on all 2,950 questions with parse failures excluded from scoring (Table~\ref{tab:main_results} counts them as wrong). \textbf{PF\%}: parse failure rate. Best per column is \textbf{bold} and second best is \underline{underlined}, computed separately above ($\geq$100B param) and below ($<$100B param) the dividing line. RGB rows receive the video frames with SAM3 overlays; tool-use rows receive no images. \fcolorbox{blue!25}{blue!6}{\strut Closed-source models in blue.} $^*$Averages over the types with at least one parseable answer (SR-3D: no parseable gap-fit answer). \fcolorbox{gray!30}{gray!15}{\strut Ours in gray.}}
\label{tab:main_results_all_std}
\tiny
\setlength{\tabcolsep}{1.5pt}
\resizebox{\textwidth}{!}{%
\begin{tabular}{@{}l|c|c|ccccc!{\color{gray!50}\vrule}c|ccccccc!{\color{gray!50}\vrule}c|ccc!{\color{gray!50}\vrule}c|c@{}}
\toprule
 & & & \multicolumn{6}{c|}{Multiple Choice (Acc $\uparrow$)} & \multicolumn{8}{c|}{Distance (MRA $\uparrow$)} & \multicolumn{4}{c|}{Volume (MRA $\uparrow$)} & \\
\cmidrule(lr){4-9} \cmidrule(lr){10-17} \cmidrule(lr){18-21}
 & \rotatebox{70}{Depth}
 & \rotatebox{70}{PF\%\,$\downarrow$}
 & \rotatebox{70}{Gap Fit}
 & \rotatebox{70}{Nearest}
 & \rotatebox{70}{Top Higher}
 & \rotatebox{70}{Wh.\ Taller}
 & \rotatebox{70}{Wh.\ Longer}
 & \rotatebox{70}{MC Avg}
 & \rotatebox{70}{How Far}
 & \rotatebox{70}{Far From Me}
 & \rotatebox{70}{How Long}
 & \rotatebox{70}{Much Taller}
 & \rotatebox{70}{Much Longer}
 & \rotatebox{70}{Fly Dist.}
 & \rotatebox{70}{Walk Dist.}
 & \rotatebox{70}{Dist Avg}
 & \rotatebox{70}{Volume}
 & \rotatebox{70}{Pour Left.}
 & \rotatebox{70}{Pour Room}
 & \rotatebox{70}{Vol Avg}
 & \rotatebox{70}{Avg} \\
\midrule
\rowcolor{blue!6} GPT-5.5 & \ding{55} & \textbf{0.0} & \textbf{80.3} & 64.0 & \textbf{91.2} & \textbf{94.2} & \underline{87.6} & \textbf{83.5} & 46.4 & 46.9 & \textbf{73.5} & \textbf{61.8} & \textbf{57.5} & 56.8 & 44.0 & \underline{55.3} & 34.0 & \underline{21.2} & \textbf{26.9} & \underline{27.4} & \textbf{59.1} \\
\rowcolor{blue!6} Gemini 2.5 Pro & \ding{55} & 2.8 & \underline{78.7} & 43.0 & 84.8 & 89.8 & 86.7 & 76.6 & 33.8 & 26.1 & 61.8 & 44.6 & 43.6 & 43.1 & 33.8 & 41.0 & 31.2 & 13.3 & 15.7 & 20.1 & 48.7 \\
\rowcolor{blue!6} Gemini 3 Flash & \ding{55} & \underline{0.1} & 72.1 & 49.5 & 82.5 & \underline{91.6} & \textbf{89.8} & 77.1 & 31.7 & 20.6 & \underline{72.3} & \underline{60.0} & \underline{53.1} & 45.2 & 40.5 & 46.2 & \textbf{42.3} & 15.6 & 24.2 & \underline{27.4} & 52.7 \\
\rowcolor{blue!6} Gemini 3.1 Pro & \ding{55} & 1.3 & 72.9 & 53.8 & \underline{88.5} & 86.2 & 84.4 & 77.2 & 27.9 & 32.3 & 65.0 & 50.2 & 44.7 & 42.9 & 39.9 & 43.3 & 27.3 & 9.7 & 18.8 & 18.6 & 49.6 \\
Qwen3-VL 235B-A22B Thinking & \ding{55} & \underline{0.1} & 75.4 & 57.0 & 80.2 & 86.2 & 81.3 & 76.0 & 40.5 & 35.5 & 51.7 & 46.1 & 39.2 & 54.9 & 40.1 & 44.0 & 27.7 & 8.6 & 16.1 & 17.4 & 49.4 \\
CuTR+Tools (Qwen3-VL 235B-A22B Thinking) & \checkmark & 1.9 & 57.4 & \textbf{78.8} & 77.0 & 83.5 & 83.1 & 75.9 & \underline{53.5} & \underline{68.5} & 37.5 & 19.8 & 19.5 & \textbf{62.0} & \underline{45.1} & 43.7 & 35.2 & 14.9 & 16.5 & 22.2 & 50.2 \\
\rowcolor{gray!15} \ours{} (Qwen3-VL 235B-A22B Thinking) & \checkmark & 2.1 & 73.8 & \underline{77.0} & 82.0 & 87.3 & 80.1 & \underline{80.0} & \textbf{55.7} & \textbf{70.8} & 62.8 & 54.5 & 41.4 & \underline{58.6} & \textbf{46.4} & \textbf{55.7} & \underline{40.6} & \textbf{22.8} & \underline{24.8} & \textbf{29.4} & \underline{58.6} \\
\midrule
Qwen3-VL 8B text-only & \ding{55} & \textbf{0.0} & 52.5 & 40.3 & 59.9 & 49.8 & 49.8 & 50.4 & 17.6 & 21.4 & 37.5 & 22.6 & 32.3 & 5.7 & 3.6 & 20.1 & 35.2 & 0.3 & 1.4 & 12.3 & 28.7 \\
Video-3D LLM & \checkmark & 55.9 & 39.0 & 56.4 & 51.7 & 48.3 & 56.2 & 50.3 & 31.8 & 45.7 & 7.5 & 25.0 & 0.0 & 6.5 & 10.1 & 18.1 & 5.9 & 11.9 & 0.0 & 5.9 & 26.4 \\
SpatialRGPT+Median & \checkmark & 20.5 & 47.5 & 27.6 & 55.6 & 68.4 & 59.8 & 51.8 & 20.8 & 28.7 & 41.4 & 11.6 & 10.2 & 0.0 & 0.0 & 16.1 & \underline{38.7} & 0.1 & 2.2 & 13.7 & 27.5 \\
SR-3D & \checkmark & 32.5 & -- & 32.6 & 50.0 & 55.3 & 57.1 & 48.8$^*$ & 22.5 & 25.4 & 10.3 & 4.8 & 4.9 & 3.6 & 4.9 & 10.9 & 7.8 & 22.9 & 16.5 & 15.7 & 22.8$^*$ \\
CuTR+Tools (Qwen3-VL 8B Instruct) & \checkmark & \underline{6.6} & 62.3 & 72.9 & 70.2 & 75.6 & \underline{82.6} & 72.7 & 55.3 & 53.0 & 35.0 & 22.7 & 23.1 & 37.7 & 20.3 & 35.3 & 31.0 & 16.2 & 16.9 & 21.4 & 45.0 \\
CuTR+Tools (Qwen3.6 35B-A3B) & \checkmark & 7.1 & 61.7 & \underline{77.7} & \textbf{77.8} & \underline{85.1} & 81.9 & \underline{76.8} & 55.4 & \underline{76.9} & 36.1 & 24.9 & 22.7 & \textbf{60.5} & \textbf{34.1} & 44.4 & 35.9 & 15.6 & 17.0 & 22.9 & 50.9 \\
\rowcolor{gray!15} \ours{} (Qwen3-VL 8B Instruct) & \checkmark & 8.4 & \underline{72.1} & 73.1 & 68.9 & 82.4 & \textbf{85.2} & 76.3 & \underline{58.0} & 54.0 & \underline{59.9} & \underline{49.6} & \underline{39.9} & 35.0 & 16.4 & \underline{44.7} & 36.9 & \underline{24.5} & \textbf{29.8} & \underline{30.4} & \underline{52.4} \\
\rowcolor{gray!15} \ours{} (Qwen3.6 35B-A3B) & \checkmark & 7.3 & \textbf{75.0} & \textbf{78.2} & \underline{73.3} & \textbf{88.7} & 80.9 & \textbf{79.2} & \textbf{60.6} & \textbf{78.5} & \textbf{65.9} & \textbf{59.0} & \textbf{49.4} & \underline{56.8} & \underline{32.6} & \textbf{57.5} & \textbf{42.0} & \textbf{26.4} & \underline{27.9} & \textbf{32.1} & \textbf{59.7} \\
\bottomrule
\end{tabular}%
}
\end{table}

\begin{table}[!t]
\centering
\caption{Main results on the 1,153 questions whose referenced objects all pass the SAM3 tracking checks (mean IoU above 0.5, correct object; tracker-pass subset), parse failures counted as wrong. \textbf{PF\%}: parse failure rate. Best per column is \textbf{bold} and second best is \underline{underlined}, computed separately above ($\geq$100B param) and below ($<$100B param) the dividing line. RGB rows receive the video frames with SAM3 overlays; tool-use rows receive no images. \fcolorbox{blue!25}{blue!6}{\strut Closed-source models in blue.} \fcolorbox{gray!30}{gray!15}{\strut Ours in gray.}}
\label{tab:main_results_tracker_pass}
\tiny
\setlength{\tabcolsep}{1.5pt}
\resizebox{\textwidth}{!}{%
\begin{tabular}{@{}l|c|c|ccccc!{\color{gray!50}\vrule}c|ccccccc!{\color{gray!50}\vrule}c|ccc!{\color{gray!50}\vrule}c|c@{}}
\toprule
 & & & \multicolumn{6}{c|}{Multiple Choice (Acc $\uparrow$)} & \multicolumn{8}{c|}{Distance (MRA $\uparrow$)} & \multicolumn{4}{c|}{Volume (MRA $\uparrow$)} & \\
\cmidrule(lr){4-9} \cmidrule(lr){10-17} \cmidrule(lr){18-21}
 & \rotatebox{70}{Depth}
 & \rotatebox{70}{PF\%\,$\downarrow$}
 & \rotatebox{70}{Gap Fit}
 & \rotatebox{70}{Nearest}
 & \rotatebox{70}{Top Higher}
 & \rotatebox{70}{Wh.\ Taller}
 & \rotatebox{70}{Wh.\ Longer}
 & \rotatebox{70}{MC Avg}
 & \rotatebox{70}{How Far}
 & \rotatebox{70}{Far From Me}
 & \rotatebox{70}{How Long}
 & \rotatebox{70}{Much Taller}
 & \rotatebox{70}{Much Longer}
 & \rotatebox{70}{Fly Dist.}
 & \rotatebox{70}{Walk Dist.}
 & \rotatebox{70}{Dist Avg}
 & \rotatebox{70}{Volume}
 & \rotatebox{70}{Pour Left.}
 & \rotatebox{70}{Pour Room}
 & \rotatebox{70}{Vol Avg}
 & \rotatebox{70}{Avg} \\
\midrule
\rowcolor{blue!6} GPT-5.5 & \ding{55} & \textbf{0.0} & \underline{92.3} & \underline{67.4} & \underline{92.2} & \textbf{97.4} & 88.5 & \underline{87.5} & 57.8 & 53.4 & \underline{75.3} & \underline{64.9} & \textbf{59.6} & 65.6 & 47.9 & \underline{60.7} & 40.6 & 25.5 & \textbf{36.0} & \underline{34.0} & \underline{64.3} \\
\rowcolor{blue!6} Gemini 2.5 Pro & \ding{55} & 2.4 & 69.2 & 52.2 & \underline{92.2} & \underline{96.1} & \underline{91.0} & 80.1 & 38.3 & 25.3 & 66.5 & 42.7 & 46.3 & 46.8 & 32.9 & 42.7 & 33.7 & 15.8 & 19.2 & 22.9 & 51.2 \\
\rowcolor{blue!6} Gemini 3 Flash & \ding{55} & \textbf{0.0} & 76.9 & 50.0 & 85.7 & 92.1 & 89.7 & 78.9 & 32.0 & 19.3 & 74.3 & 57.6 & 54.2 & 55.5 & 49.1 & 48.8 & 45.8 & 12.1 & 27.9 & 28.6 & 54.8 \\
\rowcolor{blue!6} Gemini 3.1 Pro & \ding{55} & 1.3 & 69.2 & 60.9 & \underline{92.2} & 92.1 & 88.5 & 80.6 & 29.1 & 32.9 & 68.9 & 48.9 & 50.0 & 46.6 & 46.9 & 46.2 & 33.5 & 15.1 & 19.4 & 22.7 & 53.0 \\
Qwen3-VL 235B-A22B Thinking & \ding{55} & \underline{0.1} & 76.9 & 60.9 & 80.5 & 85.5 & 87.2 & 78.2 & 44.9 & 33.8 & 57.6 & 51.7 & 39.7 & 62.1 & \underline{49.2} & 48.4 & 33.6 & 4.0 & 21.5 & 19.7 & 52.6 \\
CuTR+Tools (Qwen3-VL 235B-A22B Thinking) & \checkmark & \textbf{0.0} & 61.5 & \textbf{93.5} & 90.9 & 90.8 & \textbf{93.6} & 86.1 & \underline{82.4} & \underline{88.2} & 39.9 & 23.9 & 28.4 & \textbf{79.0} & \textbf{51.8} & 56.2 & \underline{45.9} & \underline{26.8} & 20.6 & 31.1 & 61.1 \\
\rowcolor{gray!15} \ours{} (Qwen3-VL 235B-A22B Thinking) & \checkmark & \textbf{0.0} & \textbf{100.0} & \textbf{93.5} & \textbf{96.1} & 94.7 & \textbf{93.6} & \textbf{95.6} & \textbf{89.2} & \textbf{94.8} & \textbf{80.3} & \textbf{71.1} & \underline{57.2} & \underline{72.7} & 48.8 & \textbf{73.5} & \textbf{51.7} & \textbf{41.1} & \underline{34.2} & \textbf{42.3} & \textbf{74.6} \\
\midrule
Qwen3-VL 8B text-only & \ding{55} & \textbf{0.0} & 53.8 & 39.1 & 61.0 & 50.0 & 50.0 & 50.8 & 16.4 & 22.9 & 43.7 & 26.6 & 36.8 & 6.0 & 2.1 & 22.1 & 33.5 & 0.0 & 0.8 & 11.4 & 29.5 \\
Video-3D LLM & \checkmark & 58.9 & 38.5 & 47.8 & 40.3 & 44.7 & 51.3 & 44.5 & 4.1 & 10.5 & 1.7 & 0.0 & 0.0 & 5.1 & 9.7 & 4.4 & 1.1 & 5.3 & 0.0 & 2.1 & 17.3 \\
SpatialRGPT+Median & \checkmark & 7.4 & 46.2 & 23.9 & 54.5 & 39.5 & 28.2 & 38.5 & 17.5 & 29.9 & 41.6 & 10.0 & 9.2 & 0.1 & 0.0 & 15.5 & 38.7 & 0.0 & 2.6 & 13.8 & 22.8 \\
SR-3D & \checkmark & 31.0 & 0.0 & 23.9 & 1.3 & 53.9 & 44.9 & 24.8 & 16.1 & 19.3 & 9.8 & 3.9 & 0.8 & 4.2 & 4.6 & 8.4 & 1.7 & 13.2 & 9.1 & 8.0 & 13.8 \\
CuTR+Tools (Qwen3-VL 8B Instruct) & \checkmark & 0.7 & 61.5 & \underline{84.8} & 70.1 & 81.6 & 91.0 & 77.8 & \underline{82.4} & 61.3 & 39.4 & 24.0 & 30.9 & 43.2 & 19.2 & 42.9 & 39.7 & 27.4 & 19.2 & 28.8 & 51.7 \\
CuTR+Tools (Qwen3.6 35B-A3B) & \checkmark & 0.3 & 61.5 & \textbf{93.5} & \textbf{90.9} & \underline{90.8} & 92.3 & 85.8 & \underline{82.4} & \underline{88.2} & 39.9 & 23.9 & 28.5 & \textbf{77.7} & \textbf{37.6} & 54.0 & 45.9 & 26.8 & 20.6 & 31.1 & 60.0 \\
\rowcolor{gray!15} \ours{} (Qwen3-VL 8B Instruct) & \checkmark & 0.6 & \underline{92.3} & 82.6 & 74.0 & 88.2 & \textbf{94.9} & \underline{86.4} & \textbf{89.2} & 65.5 & \underline{79.1} & \underline{68.7} & \underline{51.5} & 38.7 & 12.2 & \underline{57.8} & \underline{48.3} & \underline{40.4} & \textbf{35.1} & \underline{41.2} & \underline{64.0} \\
\rowcolor{gray!15} \ours{} (Qwen3.6 35B-A3B) & \checkmark & \underline{0.1} & \textbf{100.0} & \textbf{93.5} & \underline{87.0} & \textbf{94.7} & \underline{93.6} & \textbf{93.8} & \textbf{89.2} & \textbf{94.8} & \textbf{80.5} & \textbf{71.1} & \textbf{56.8} & \underline{71.4} & \underline{32.4} & \textbf{70.9} & \textbf{51.6} & \textbf{44.5} & \underline{34.3} & \textbf{43.5} & \textbf{73.0} \\
\bottomrule
\end{tabular}%
}
\end{table}

\begin{table}[!t]
\centering
\caption{Main results on the 1,797 questions with at least one referenced object that fails these checks (tracker-failure subset), parse failures counted as wrong. \textbf{PF\%}: parse failure rate. Best per column is \textbf{bold} and second best is \underline{underlined}, computed separately above ($\geq$100B param) and below ($<$100B param) the dividing line. RGB rows receive the video frames with SAM3 overlays; tool-use rows receive no images. \fcolorbox{blue!25}{blue!6}{\strut Closed-source models in blue.} \fcolorbox{gray!30}{gray!15}{\strut Ours in gray.}}
\label{tab:main_results_tracker_failure}
\tiny
\setlength{\tabcolsep}{1.5pt}
\resizebox{\textwidth}{!}{%
\begin{tabular}{@{}l|c|c|ccccc!{\color{gray!50}\vrule}c|ccccccc!{\color{gray!50}\vrule}c|ccc!{\color{gray!50}\vrule}c|c@{}}
\toprule
 & & & \multicolumn{6}{c|}{Multiple Choice (Acc $\uparrow$)} & \multicolumn{8}{c|}{Distance (MRA $\uparrow$)} & \multicolumn{4}{c|}{Volume (MRA $\uparrow$)} & \\
\cmidrule(lr){4-9} \cmidrule(lr){10-17} \cmidrule(lr){18-21}
 & \rotatebox{70}{Depth}
 & \rotatebox{70}{PF\%\,$\downarrow$}
 & \rotatebox{70}{Gap Fit}
 & \rotatebox{70}{Nearest}
 & \rotatebox{70}{Top Higher}
 & \rotatebox{70}{Wh.\ Taller}
 & \rotatebox{70}{Wh.\ Longer}
 & \rotatebox{70}{MC Avg}
 & \rotatebox{70}{How Far}
 & \rotatebox{70}{Far From Me}
 & \rotatebox{70}{How Long}
 & \rotatebox{70}{Much Taller}
 & \rotatebox{70}{Much Longer}
 & \rotatebox{70}{Fly Dist.}
 & \rotatebox{70}{Walk Dist.}
 & \rotatebox{70}{Dist Avg}
 & \rotatebox{70}{Volume}
 & \rotatebox{70}{Pour Left.}
 & \rotatebox{70}{Pour Room}
 & \rotatebox{70}{Vol Avg}
 & \rotatebox{70}{Avg} \\
\midrule
\rowcolor{blue!6} GPT-5.5 & \ding{55} & \textbf{0.0} & \underline{77.1} & 62.9 & \textbf{90.7} & \textbf{92.6} & \underline{87.1} & \textbf{82.1} & \textbf{40.6} & \underline{37.8} & \textbf{70.9} & \underline{60.0} & \textbf{56.3} & \underline{52.3} & \underline{41.9} & \textbf{51.4} & 22.8 & \textbf{18.8} & \textbf{21.8} & \underline{21.1} & \textbf{55.6} \\
\rowcolor{blue!6} Gemini 2.5 Pro & \ding{55} & 3.1 & \textbf{81.2} & 40.0 & 80.7 & 86.6 & 84.4 & 74.6 & 31.0 & 26.9 & 54.7 & 44.4 & 41.6 & 37.7 & 23.9 & 37.2 & \underline{25.4} & 10.7 & 12.9 & 16.3 & 45.5 \\
\rowcolor{blue!6} Gemini 3 Flash & \ding{55} & 0.2 & 70.8 & 49.3 & 80.7 & \underline{91.3} & \textbf{89.8} & \underline{76.4} & 31.5 & 22.3 & \underline{69.5} & \textbf{61.3} & \underline{52.1} & 39.5 & 35.7 & \underline{44.6} & \textbf{36.2} & \underline{17.7} & \textbf{21.8} & \textbf{25.2} & \underline{51.3} \\
\rowcolor{blue!6} Gemini 3.1 Pro & \ding{55} & 1.2 & 70.8 & 51.4 & \underline{86.4} & 83.2 & 82.3 & 74.8 & 26.3 & 30.5 & 59.5 & 48.2 & 41.3 & 40.7 & 33.9 & 40.1 & 15.4 & 6.5 & 17.8 & 13.2 & 46.3 \\
Qwen3-VL 235B-A22B Thinking & \ding{55} & \underline{0.1} & 75.0 & 55.7 & 80.0 & 86.6 & 78.2 & 75.1 & 38.3 & \underline{37.8} & 43.6 & 42.9 & 38.8 & 51.2 & 35.2 & 41.1 & 16.7 & 11.2 & 13.0 & 13.6 & 46.9 \\
CuTR+Tools (Qwen3-VL 235B-A22B Thinking) & \checkmark & 3.1 & 56.2 & \textbf{70.0} & 69.3 & 62.4 & 67.3 & 65.1 & \underline{38.9} & \textbf{41.0} & 34.1 & 17.4 & 14.8 & \textbf{53.2} & 41.6 & 34.4 & 17.2 & 8.2 & 14.3 & 13.2 & 40.4 \\
\rowcolor{gray!15} \ours{} (Qwen3-VL 235B-A22B Thinking) & \checkmark & 3.4 & 66.7 & \underline{67.1} & 74.3 & 67.1 & 59.9 & 67.0 & 38.7 & 36.7 & 38.7 & 44.9 & 32.9 & 51.3 & \textbf{45.0} & 41.2 & 22.0 & 12.4 & \underline{19.5} & 18.0 & 45.1 \\
\midrule
Qwen3-VL 8B text-only & \ding{55} & \textbf{0.0} & 52.1 & 40.7 & \underline{59.3} & 49.7 & 49.7 & 50.3 & 18.2 & 19.4 & 28.8 & 20.3 & 29.9 & 5.5 & 4.4 & 18.1 & \textbf{38.2} & 0.5 & 1.8 & 13.5 & 27.9 \\
Video-3D LLM & \checkmark & 54.0 & 37.5 & 50.0 & 32.1 & 34.2 & 46.9 & 40.2 & 5.1 & 13.5 & 0.5 & 0.7 & 0.0 & 4.0 & 6.6 & 4.4 & 2.6 & 2.4 & 0.0 & 1.7 & 15.8 \\
SpatialRGPT+Median & \checkmark & 28.9 & 45.8 & 22.1 & 48.6 & 33.6 & 34.7 & 37.0 & 19.5 & 3.2 & 9.3 & 11.1 & 9.7 & 0.0 & 0.0 & 7.5 & 0.0 & 0.1 & 1.5 & 0.5 & 15.9 \\
SR-3D & \checkmark & 33.4 & 0.0 & 25.0 & 0.7 & 43.0 & 49.7 & 23.7 & 17.1 & 18.2 & 6.2 & 2.3 & 3.6 & 3.0 & 4.1 & 7.8 & 5.2 & \textbf{19.9} & 14.4 & 13.2 & 14.2 \\
CuTR+Tools (Qwen3-VL 8B Instruct) & \checkmark & \underline{10.4} & \underline{62.5} & 52.9 & 57.1 & 49.7 & 55.1 & 55.5 & \underline{41.6} & 39.3 & 27.4 & 20.6 & 16.9 & 30.7 & 20.5 & 28.1 & 16.4 & 9.8 & 15.0 & 13.7 & 34.4 \\
CuTR+Tools (Qwen3.6 35B-A3B) & \checkmark & 11.4 & 60.4 & \underline{68.6} & \textbf{65.0} & \underline{69.1} & \textbf{68.0} & \textbf{66.2} & 36.9 & \textbf{46.3} & 25.6 & 19.9 & 15.2 & \textbf{44.8} & \underline{29.7} & 31.2 & 15.6 & 8.2 & 14.1 & 12.6 & \underline{39.2} \\
\rowcolor{gray!15} \ours{} (Qwen3-VL 8B Instruct) & \checkmark & 13.5 & \textbf{66.7} & 54.3 & 52.9 & 55.7 & 55.1 & 56.9 & \textbf{42.1} & 35.7 & \underline{30.9} & \underline{36.1} & \underline{30.6} & 29.3 & 18.0 & \underline{31.8} & 17.2 & 8.0 & \underline{18.0} & \underline{14.4} & 36.7 \\
\rowcolor{gray!15} \ours{} (Qwen3.6 35B-A3B) & \checkmark & 11.9 & \textbf{66.7} & \textbf{69.3} & 57.9 & \textbf{72.5} & \underline{62.6} & \underline{65.8} & 40.8 & \underline{43.3} & \textbf{36.8} & \textbf{42.8} & \textbf{34.7} & \underline{44.3} & \textbf{30.2} & \textbf{39.0} & \underline{20.3} & \underline{13.6} & \textbf{20.7} & \textbf{18.2} & \textbf{43.8} \\
\bottomrule
\end{tabular}%
}
\end{table}

%----------------------------------------------------------------------
\section{Evaluating Scene Quality}
\label{app:scene_quality}
We evaluate the quality of our reconstructed scene here (Table~\ref{tab:scene_quality}) by looking at the error of reconstructed objects that the tracker found (excluding undetected objects). Both methods start from the same SAM3 tracks. \ours{}'s boxes are far more accurate than CuTR's on typical objects: median IoU 0.58 vs.\ 0.22, height error 7.2\% vs.\ 37.4\%, longest-dimension error 5.9\% vs.\ 33.0\%, and center error 4.4 vs.\ 6.8\,cm (90th percentile 24.6 vs.\ 89.1\,cm); CuTR's yaw is slightly more accurate (median 6.7$^\circ$ vs.\ 8.7$^\circ$). \ours{}'s functional volume has the lowest median and 90th-percentile error, and its median distance errors are less than half of CuTR's.

\begin{table}[h]
\centering
\small
\caption{Accuracy of the constructed scene against ADT ground truth, over objects that SAM3 tracks. Box and distance rows: the 1,447 tracked objects of the 2,092 referenced by all questions (the other 645 are never tracked and have no reconstruction). Functional-volume rows: the 163 tracked objects of the 265 referenced by the volume and pour questions (the other 102 are never tracked).}
\label{tab:scene_quality}
\resizebox{\textwidth}{!}{%
\begin{tabular}{@{}llrrr@{}}
\toprule
Quantity & Method & P10 & P50 & P90 \\
\midrule
3D IoU & \ours{} & 0.16 & 0.58 & 0.77 \\
3D IoU & CuTR & 0.00 & 0.22 & 0.41 \\
\midrule
Center error, cm & \ours{} & 0.7 & 4.4 & 24.6 \\
Center error, cm & CuTR & 2.7 & 6.8 & 89.1 \\
\midrule
Yaw error, deg & \ours{} & 0.7 & 8.7 & 35.2 \\
Yaw error, deg & CuTR & 1.2 & 6.7 & 30.9 \\
\midrule
Height error, \% & \ours{} & 1.2 & 7.2 & 33.2 \\
Height error, \% & CuTR & 17.2 & 37.4 & 71.9 \\
\midrule
Longest-dimension error, \% & \ours{} & 0.8 & 5.9 & 43.2 \\
Longest-dimension error, \% & CuTR & 10.9 & 33.0 & 62.6 \\
\midrule
Box-volume error, \% & \ours{} & 6.4 & 31.1 & 179.4 \\
Box-volume error, \% & CuTR & 27.5 & 65.2 & 95.7 \\
\midrule
Functional volume err., \% & \ours{} (SAM3D, rescaled to \ours{} box) & 5.7 & 23.2 & 65.9 \\
Functional volume err., \% & CuTR box volume & 5.7 & 33.4 & 182.8 \\
Functional volume err., \% & CuTR + SAM3D (rescaled to CuTR box) & 22.8 & 58.1 & 84.0 \\
\midrule
Object-to-object distance err., \% & \ours{} & 0.4 & 2.5 & 27.7 \\
Object-to-object distance err., \% & CuTR & 1.5 & 5.9 & 28.2 \\
\midrule
Distance-from-camera err., \% & \ours{} & 0.1 & 1.1 & 11.3 \\
Distance-from-camera err., \% & CuTR & 0.3 & 3.0 & 17.1 \\
\bottomrule
\end{tabular}%
}
\end{table}

%----------------------------------------------------------------------
\clearpage
\section{R3D Scene Construction Method Ablation}
\label{app:scene_ablation}
Here we explore the importance of the details of our scene construction method. Table~\ref{tab:scene_ablation_qa} reports question-answering accuracy when components of the scene construction are removed or replaced, and Table~\ref{tab:scene_ablation_boxes} reports the accuracy of the resulting boxes and volumes against ADT ground truth. We find that combining all components of our method provides the best results.

\begin{table}[h]
\centering
\small
\caption{Question-answering ablation of the scene construction (\ours{} + Qwen3.6 35B-A3B, parse failures counted as wrong). ``First frame'': the box from the first frame in which SAM3 detects the object with at least 10 depth points; ``Median'': each frame's points are lifted on their own and the box parameters are the element-wise median over frames; ``Fused'': one box over the points lifted from all frames. ``Vote'': multiview voting; ``KNN'': KNN outlier filtering of the point cloud. Volume comes from the box itself (Box) or from the SAM3D mesh rescaled to the box (SAM3D).}
\label{tab:scene_ablation_qa}
\begin{tabular}{@{}lllllrrr@{}}
\toprule
Boxes & Vote & KNN & Volume & Overall & MC & Dist & Vol \\
\midrule
First frame & -- & -- & Box & 46.2 & 73.4 & 43.8 & 6.4 \\
Median & -- & -- & Box & 45.1 & 72.2 & 43.9 & 2.7 \\
Fused & -- & -- & Box & 31.5 & 64.6 & 21.1 & 0.6 \\
Fused & \checkmark & -- & Box & 43.5 & 69.9 & 41.9 & 2.9 \\
Fused & -- & \checkmark & Box & 48.4 & 73.7 & 49.2 & 4.5 \\
Fused & \checkmark & \checkmark & Box & 51.3 & \textbf{74.1} & \textbf{53.2} & 8.8 \\
\rowcolor{gray!15} Fused & \checkmark & \checkmark & SAM3D & \textbf{55.6} & \textbf{74.1} & \textbf{53.2} & \textbf{30.1} \\
\bottomrule
\end{tabular}
\end{table}

\begin{table}[h]
\centering
\small
\caption{Accuracy of each ablation scene in Table~\ref{tab:scene_ablation_qa} against ADT ground truth: 3D IoU of the oriented boxes, center error, relative error of height and longest dimension, and relative error of the volume against the ground-truth capacity over the 163 objects of the volume and pour questions that SAM3 tracks (the box volume for Box rows, the rescaled SAM3D functional volume for the SAM3D row). These questions reference 265 objects; the other 102 are never tracked by SAM3 and have no reconstruction.}
\label{tab:scene_ablation_boxes}
\setlength{\tabcolsep}{4pt}
\begin{tabular}{@{}llllrrrrr@{}}
\toprule
Boxes & Vote & KNN & Volume & 3D IoU $\uparrow$ & Center (cm) $\downarrow$ & Height $\downarrow$ & Longest dim.\ $\downarrow$ & Volume $\downarrow$ \\
\midrule
\multicolumn{9}{@{}l}{\textit{Median over objects}} \\
First frame & -- & -- & Box & 0.35 & 13.2 & 7.1\% & 43.7\% & 360.0\% \\
Median & -- & -- & Box & 0.37 & 12.1 & \textbf{6.3\%} & 34.6\% & 491.2\% \\
Fused & -- & -- & Box & 0.03 & 61.9 & 68.0\% & 315.7\% & 244998.9\% \\
Fused & \checkmark & -- & Box & 0.32 & 12.4 & 6.9\% & 39.8\% & 384.7\% \\
Fused & -- & \checkmark & Box & 0.54 & 5.4 & \textbf{6.3\%} & 7.9\% & 164.4\% \\
Fused & \checkmark & \checkmark & Box & \textbf{0.58} & \textbf{4.4} & 7.2\% & \textbf{5.9\%} & 114.1\% \\
\rowcolor{gray!15} Fused & \checkmark & \checkmark & SAM3D & \textbf{0.58} & \textbf{4.4} & 7.2\% & \textbf{5.9\%} & \textbf{23.2\%} \\
\midrule
\multicolumn{9}{@{}l}{\textit{Mean over objects}} \\
First frame & -- & -- & Box & 0.35 & 31.4 & 110.1\% & 122.7\% & 4570.2\% \\
Median & -- & -- & Box & 0.36 & 30.2 & 111.5\% & 113.0\% & 5122.8\% \\
Fused & -- & -- & Box & 0.11 & 80.7 & 545.5\% & 561.4\% & 550143.3\% \\
Fused & \checkmark & -- & Box & 0.33 & 39.1 & 133.9\% & 156.3\% & 8606.3\% \\
Fused & -- & \checkmark & Box & 0.49 & 23.6 & 70.9\% & 84.8\% & 5077.9\% \\
Fused & \checkmark & \checkmark & Box & \textbf{0.53} & \textbf{18.1} & \textbf{34.9\%} & \textbf{56.0\%} & 685.9\% \\
\rowcolor{gray!15} Fused & \checkmark & \checkmark & SAM3D & \textbf{0.53} & \textbf{18.1} & \textbf{34.9\%} & \textbf{56.0\%} & \textbf{363.5\%} \\
\bottomrule
\end{tabular}
\end{table}

%----------------------------------------------------------------------
\clearpage
\section{Text-Based Prompts In Place of Tool Calling}
\label{app:tool_ablation}
Our main motivation for using tools as opposed to text-based representations is to avoid flooding the model context with a potentially large set of text tokens, instead allowing the model to inspect the scene graph through tools. A general-purpose model needs to know when to include the scene as context, and when not to (for example, when answering general-knowledge questions). Nevertheless, adding the scene graph as text is another viable option. We evaluate the performance of using text instead of tools here.

In Table~\ref{tab:text_vs_tools} we explore the performance of using a text-based representation. We serialize the same reconstructed scene the tools read into the prompt---every object's ID, name, gravity-aligned box center, size and yaw, its SAM3D functional volume, and the camera position at question time---and query the same LLM (Qwen3.6 35B-A3B) without tools. Objects missing from the scene are missing from the table, exactly as they are missing from the tools. 

The two interfaces perform similarly overall (55.6\% for tools and 56.6\% for text with parse failures counted as wrong; 59.7\% and 59.5\% with parse failures excluded). The tool calling approach outperforms text on questions that require closest-point distance between two bounding boxes (How Far) or the user and a bounding box (Far From Me), and the text approach outperforms on questions that ask for distances that include object centers (Walk, Fly). On the latter, the reasoning model typically fails to account for this, though it has the necessary data to do so (it has access to bounding box centers), even when the tool descriptions explicitly mention that distances use closest-point instead of object-center. The text table costs a median of 2,420 prompt tokens per question and grows linearly with the number of objects, whereas tool outputs do not scale with the number of objects with the exception of the object listing function.

\begin{table}[h]
\centering
\scriptsize
\setlength{\tabcolsep}{2.5pt}
\caption{Tool calling vs.\ supplying the same reconstructed scene as a text table (object IDs, names, gravity-aligned box centers, sizes and yaw, SAM3D functional volumes, camera position) to the same LLM without tools. Qwen3.6 35B-A3B, all 2,950 questions, parse failures counted as wrong; the last column excludes them. Objects missing from the scene are missing from both interfaces.}
\label{tab:text_vs_tools}
\resizebox{\textwidth}{!}{%
\begin{tabular}{@{}lrrrrrrrrrrrrrrrrr@{}}
\toprule
 & \rotatebox{70}{Gap Fit} & \rotatebox{70}{Nearest} & \rotatebox{70}{Top Higher} & \rotatebox{70}{Wh. Taller} & \rotatebox{70}{Wh. Longer} & \rotatebox{70}{How Far} & \rotatebox{70}{Far From Me} & \rotatebox{70}{How Long} & \rotatebox{70}{Much Taller} & \rotatebox{70}{Much Longer} & \rotatebox{70}{Fly Dist.} & \rotatebox{70}{Walk Dist.} & \rotatebox{70}{Volume} & \rotatebox{70}{Pour Left.} & \rotatebox{70}{Pour Room} & \rotatebox{70}{Avg} & \rotatebox{70}{Avg (PF excl.)} \\
\midrule
Tool calling (\ours{}) & \textbf{73.8} & 75.3 & 68.2 & \textbf{80.0} & \textbf{73.3} & \textbf{57.1} & \textbf{73.3} & \textbf{62.1} & 53.2 & 42.4 & 53.5 & 31.0 & \textbf{39.9} & 24.8 & 25.6 & 55.6 & \textbf{59.7} \\
Same scene as text, no tools & 67.2 & \textbf{77.4} & \textbf{77.4} & 76.4 & 72.9 & 42.6 & 60.5 & 58.2 & \textbf{54.8} & \textbf{42.7} & \textbf{70.3} & \textbf{58.2} & 37.1 & \textbf{25.2} & \textbf{27.9} & \textbf{56.6} & 59.5 \\
\bottomrule
\end{tabular}%
}
\end{table}

%----------------------------------------------------------------------
\section{Evaluations With Ground Truth Scenes and Tools}
\label{app:gt_evals}
We measure how much of \ours{}'s error comes from the scene and how much from the LLM's tool use and reasoning (Table~\ref{tab:gt_evals}). The ``GT scene'' refers to using ground truth labels in place of the R3D scene. ``GT tools'' refers to plugging in the scene into tools that yield the exact computation required. (Using GT for both scene and tools is slightly below 100\%, due to tool output rounding to the nearest centimeter.) % NOTE: leave this wording of the parenthetical alone.

\begin{table}[h]
\centering
\small
\caption{Upper-bound evaluations on all 2,950 questions (parse failures counted as wrong). \emph{Scene}: \ours{}'s reconstructed scene, or the ground-truth scene, in which every referenced object is present with its ADT box and functional volume, and the camera pose is ground truth. \emph{Tools}: with \ours{} tools, an LLM agent answers using the tools of Section~\ref{sec:tool_calling}; with GT tools, a fixed formula per question type, given the question's objects and parameters directly, computes the answer from the scene without an LLM, and abstains when a referenced object is missing from the scene. In addition to overall results, we show results for questions in which all objects are in the scene, and at least one object is missing from the scene (due to tracker failure).}
\label{tab:gt_evals}
\begin{tabular}{@{}lllrrr@{}}
\toprule
& & & & \multicolumn{2}{c}{Objects in scene} \\
\cmidrule(l){5-6}
Scene & Tools & Reasoner & Overall & All in scene & $\geq$1 missing \\
\midrule
\ours{} & \ours{} & Qwen3-VL 8B Instruct & 47.6 & 63.3 & 32.6 \\
 & & Qwen3.6 35B-A3B & 55.6 & 72.4 & 39.2 \\
 & & Qwen3-VL 235B-A22B Thinking & 57.0 & 73.9 & 40.5 \\
\midrule
\ours{} & GT & none (fixed formula) & 37.4 & 79.0 & 0.0 \\
\midrule
GT & \ours{} & Qwen3-VL 8B Instruct & 80.0 & 80.0 & -- \\
 & & Qwen3.6 35B-A3B & 93.3 & 93.3 & -- \\
 & & Qwen3-VL 235B-A22B Thinking & 94.8 & 94.8 & -- \\
\midrule
GT & GT & none (fixed formula) & 99.6 & 99.6 & -- \\
\bottomrule
\end{tabular}
\end{table}

%----------------------------------------------------------------------
\section{SAM3D Reconstruction Quality And Ablations}
\label{app:recon}

\begin{table}[!ht]
\centering
\caption{Chamfer distance (cm$^2$) between 3D representations and ground truth meshes, reported as 5th, 50th, and 95th percentiles over the objects of the Volume questions (pour questions excluded) that SAM3 tracks: 103 of the 164 objects these questions reference; the other 61 are never tracked by SAM3 and have no reconstruction. \colorbox{gray!15}{Gray row}: oracle using ground-truth segmentation masks. Bold: best among rows without ground-truth masks.}
\label{tab:cd_comparison}
\small
\setlength{\tabcolsep}{6pt}
\begin{tabular}{@{}l|rrr|rrr|rrr@{}}
\toprule
 & \multicolumn{3}{c|}{CD: Pred$\to$GT} & \multicolumn{3}{c|}{CD: GT$\to$Pred} & \multicolumn{3}{c}{CD (Full)} \\
\cmidrule(lr){2-4} \cmidrule(lr){5-7} \cmidrule(lr){8-10}
 & P5 & P50 & P95 & P5 & P50 & P95 & P5 & P50 & P95 \\
\midrule

R3D Pointcloud & 0.15 & 1.58 & 17.5 & 0.56 & 3.20 & 27.5 & 0.80 & 4.89 & 42.6 \\
R3D Aniso & 0.16 & 3.11 & 133.2 & 0.19 & 1.72 & 28.3 & 0.32 & 5.01 & 193.7 \\
R3D Iso & \textbf{0.12} & \textbf{0.49} & \textbf{4.89} & \textbf{0.14} & \textbf{0.67} & \textbf{5.47} & \textbf{0.28} & \textbf{1.18} & \textbf{11.3} \\
\rowcolor{gray!15} R3D Iso (GT Seg) & 0.13 & 0.59 & 5.27 & 0.14 & 0.69 & 5.87 & 0.29 & 1.22 & 10.2 \\
\bottomrule
\end{tabular}
\end{table}

Here we examine the quality of the 3D representations described in Section~\ref{sec:volumetric} using Chamfer distance (CD). Given a predicted point set $\mathcal{P}$ and a ground truth point set $\mathcal{G}$, we compute two directional distances:
\begin{align}
\text{CD}_{\text{P}\to\text{G}} &= \frac{1}{|\mathcal{P}|}\sum_{\mathbf{p}\in\mathcal{P}} \min_{\mathbf{g}\in\mathcal{G}} \|\mathbf{p}-\mathbf{g}\|^2, \label{eq:cd_p2g} \\
\text{CD}_{\text{G}\to\text{P}} &= \frac{1}{|\mathcal{G}|}\sum_{\mathbf{g}\in\mathcal{G}} \min_{\mathbf{p}\in\mathcal{P}} \|\mathbf{g}-\mathbf{p}\|^2. \label{eq:cd_g2p}
\end{align}
$\text{CD}_{\text{P}\to\text{G}}$ measures \emph{precision} (how close predicted points are to the true surface), while $\text{CD}_{\text{G}\to\text{P}}$ measures \emph{completeness} (how much of the true surface is covered). The full CD is their sum. We compute CD between each representation and the ground truth mesh for the 103 objects of the Volume question type that SAM3 tracks (the other 61 are never tracked and have no mesh) by sampling 5000 points from ground truth and predicted meshes. Table~\ref{tab:cd_comparison} shows the results. ``R3D Iso'' refers to our method, and ``R3D Aniso'' refers to an ablation in which each bounding box axis is scaled independently:
\begin{equation}
s_i = \frac{d_i^{\text{bbox}}}{d_i^{\text{mesh}}}, \quad i \in \{1, 2, 3\},
\label{eq:aniso_scale}
\end{equation}
where $d_i^{\text{bbox}}$ and $d_i^{\text{mesh}}$ are the extents of the predicted bounding box and mesh bounding box along each principal axis, respectively. R3D Iso (GT Seg) refers to an ablation in which ground truth segmentations are used as the input to SAM3D instead of SAM3 masks.

The R3D Iso mesh achieves the strongest results among methods that don't use ground truth masks. Compared to the point cloud, R3D Iso achieves much stronger completeness ($\text{CD}_{\text{G}\to\text{P}}$), filling in surface regions unseen by the point cloud. R3D Iso (GT Seg) gives similar results (better only on the 95th percentile of the full CD), so the SAM3 masks are not the main source of reconstruction error. For fair comparison, we do not allow R3D to ingest ground truth masks in our main results in Table~\ref{tab:main_results}.

\begin{table}[h]
\centering
\small
\caption{Volume from rescaled SAM3D meshes (ours) vs.\ from the depth-lifted bounding box, everything else fixed (\ours{} + Qwen3.6 35B-A3B, all 2,950 questions, parse failures counted as wrong). The 12 non-volume types are identical because only \texttt{get\_object\_volume} changes.}
\label{tab:sam3d_vs_box}
\begin{tabular}{@{}llrr@{}}
\toprule
Category & Type & Box volume & SAM3D (ours) \\
\midrule
Multiple choice & Gap Fit & 73.8 & 73.8 \\
 & Nearest & 75.3 & 75.3 \\
 & Top Higher & 68.2 & 68.2 \\
 & Which Taller & 80.0 & 80.0 \\
 & Which Longer & 73.3 & 73.3 \\
 & \textit{Average} & 74.1 & 74.1 \\
\midrule
Distance & How Far & 57.1 & 57.1 \\
 & Far From Me & 73.3 & 73.3 \\
 & How Long & 62.1 & 62.1 \\
 & Much Taller & 53.2 & 53.2 \\
 & Much Longer & 42.4 & 42.4 \\
 & Fly Dist. & 53.5 & 53.5 \\
 & Walk Dist. & 31.0 & 31.0 \\
 & \textit{Average} & 53.2 & 53.2 \\
\midrule
Volume & Volume & 14.0 & \textbf{39.9} \\
 & Pour Left. & 6.5 & \textbf{24.8} \\
 & Pour Room & 6.1 & \textbf{25.6} \\
 & \textit{Average} & 8.8 & \textbf{30.1} \\
\midrule
\multicolumn{2}{@{}l}{Overall (parse failures counted as wrong)} & 51.3 & \textbf{55.6} \\
\multicolumn{2}{@{}l}{Overall (parse failures excluded)} & 55.1 & \textbf{59.7} \\
\bottomrule
\end{tabular}
\end{table}

Table~\ref{tab:sam3d_vs_box} isolates the effect of SAM3D on question answering. Replacing the rescaled SAM3D volume with the volume of the depth-lifted bounding box leaves every non-volume answer unchanged and lowers the volume average from 30.1 to 8.8. A bounding box counts the air around and inside an object, so it overestimates capacity; the mesh's cavity removes that air.

%----------------------------------------------------------------------
%----------------------------------------------------------------------
%----------------------------------------------------------------------
\end{document}